\documentclass{article}

    \PassOptionsToPackage{numbers}{natbib}

 \usepackage[eandd, final]{neurips_2026}

\usepackage[utf8]{inputenc} %
\usepackage[T1]{fontenc}    %
\usepackage{hyperref}       %
\usepackage{url}            %
\usepackage{booktabs}       %
\usepackage{amsfonts}       %
\usepackage{nicefrac}       %
\usepackage{microtype}      %
\usepackage{xcolor}         %
\usepackage{amsmath}

\hypersetup{
  hidelinks,
  colorlinks=false,      %
  allcolors=[HTML]{4477AA}    %
}

\usepackage{graphicx}
\usepackage[nolist]{acronym}
\usepackage{subcaption}
\usepackage{cleveref}
\usepackage{multirow}
\usepackage{xurl}
\usepackage{wrapfig}
\usepackage{adjustbox}

\newlength{\smalltabcolsep}
\newcommand{\datasetcaption}[2]{%
    #1~(\includegraphics[height=5pt]{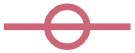})\,/\,\allowbreak#2~(\includegraphics[height=5pt]{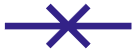})%
}

\title{BIAS-ID: A Framework for Analyzing Transformation Biases in AI-Generated Image Detectors}

\author{%
  Jonas Ricker\\
  Ruhr University Bochum\\
  Bochum, Germany\\
  \And
  Asja Fischer\\
  Ruhr University Bochum\\
  Bochum, Germany\\
  \And
  Erwin Quiring\\
  \_fbeta\\
  Berlin, Germany\\
}

\begin{document}

\maketitle

\begin{acronym}
    \acro{GenAI}{generative artificial intelligence}
    \acro{AIGC}{AI-generated content}
    \acro{AIGI}{AI-generated image}
    \acro{DM}{diffusion model}
    \acro{GAN}{generative adversarial network}
    \acro{LDM}{latent diffusion model}
    \acro{VAE}{variational autoencoder}
\end{acronym}

\begin{abstract}

Given the surge of harmful AI-generated imagery online, reliably distinguishing authentic images from generated ones has become an urgent research topic. While many proposed detection methods demonstrate high accuracy in controlled settings, their performance often degrades substantially when tested on real-world data. A potential root cause is subtle bias in the detectors' training data. As a result, detectors may rely on spurious correlations instead of learning true forensic artifacts. While recent studies have identified this problem and developed countermeasures, there is not yet an established protocol to evaluate how biased a detector actually is. In this work, we therefore take a step back: First, we discuss what it means for a detector to be biased, and how this differs from a lack of robustness. Second, we propose BIAS-ID, a transparent framework for analyzing and quantifying the presence of transformation biases in AI-generated image detectors. We validate our framework by evaluating six detectors across two datasets, revealing that several state-of-the-art detection methods are strongly affected by common, processing-related biases. Our results highlight the importance of bias-aware evaluation for developing reliable AI-generated image detectors. We provide an easy-to-use library and code to reproduce our results at\thickspace \href{https://github.com/jonasricker/biasid}{\texttt{github.com/jonasricker/biasid}}.

\end{abstract}

\section{Introduction}
\label{sec:intro}

\Ac{AIGC} has become so deceptively authentic that humans are often unable to distinguish real from generated media.
Especially for \acp{AIGI}, several studies have shown that people can no longer reliably tell apart real from generated images~\cite{nightingaleAIsynthesizedFacesAre2022,Frank2024RepresentativeStudy,cookeGoodCoinToss2025}, opening the door to disinformation~\cite{fbiMaliciousActorsAlmost2021,dufourAMMeBaLargescaleSurvey2024}, scamming~\cite{fbiCriminalsUseGenerative2024,rickerAIgeneratedFacesReal2024}, and a general erosion of trust in visual media~\cite{ryan-mosleyHowGenerativeAI,gretelkahnWillAIgeneratedImages2023}.

Therefore, developing reliable detection methods for \acp{AIGI} is an urgent and highly active area of research.
In contrast to proactive approaches, such as signatures~\cite{c2pa} and watermarks~\cite{Zhu2018hidden,fernandezStableSignatureRooting2023,Gunn2025Undetectable}, 
passive detection approaches do not rely on intentional modifications during image generation. 
Instead, they analyze the final image itself to detect traces or artifacts indicating it was generated by AI.
Here, data-driven approaches, trained on large quantities of real and generated images, have recently achieved impressive results~\cite{wangCNNgeneratedImagesAre2020,frankLeveragingFrequencyAnalysis2020,ojhaUniversalFakeImage2023,chenDRCTDiffusionReconstruction2024,koutlisLeveragingRepresentationsIntermediate2025,yanSanityCheckAIgenerated2024a,karageorgiouAnyResolutionAIGeneratedImage2025,guillaroBiasfreeTrainingParadigm2025}.

However, it has been repeatedly observed that their performance can drastically decrease when tested on real-world data~\cite{gragnanielloAreGANGenerated2021,yanSanityCheckAIgenerated2024a,guillaroBiasfreeTrainingParadigm2025}.
Besides a general lack of generalization, recent work suggests that subtle \emph{biases} in the detector's training data may be responsible for these deviations~\cite{grommeltFakeJPEGRevealing2024,rajanAlignedDatasetsImprove2025,zhengBreakingSemanticArtifacts2024,gyeReducingContentBias2025,guillaroBiasfreeTrainingParadigm2025,lorchLandscapeMoreSecure2024}.
These can arise from image transformations, such as compression, resizing, or rotation, as well as from inherent image properties, including content, visual quality, and complexity.
When these characteristics are distributed unevenly between real and fake training data, a detector may learn \emph{spurious correlations}.
For instance, if it is trained on compressed real images and uncompressed fake images, the detector may incorrectly treat compression artifacts as a sign that an image is real. 
Thus, it may misclassify uncompressed real images as fake, as well as compressed fake images as real.
Such biases also make it easy for malicious actors to evade \ac{AIGI} detectors by simply applying (often imperceptible) transformations to an image.

In this work, we introduce BIAS-ID (\textbf{b}ias \textbf{i}nspection and \textbf{a}nalysis for \textbf{s}ynthetic \textbf{i}mage \textbf{d}etectors), a comprehensive framework for analyzing the presence of transformation biases in \ac{AIGI} detectors.
In contrast to prior work, which focuses on how biases affect the \emph{detection performance}, we argue that directly analyzing a detector's \emph{predictions} provides richer insights. 
In particular, we propose analyzing the \emph{score shift}, the amount by which a transformation changes an image's predicted probability of being AI-generated. This metric reveals to what extent and toward which class (real or fake) a detector is biased regarding a transformation.
By aggregating score shifts from real and generated images and analyzing the sensitivity to a transformation at different levels, we obtain the \emph{aggregated transformation sensitivity (ATS)}, which serves as a scalar metric to compare different detectors w.r.t.\ how biased they are.
This analysis provides researchers with actionable information: it helps identify which transformations systematically skew detector behavior and can thus reveal potential artifacts or imbalances in the training data. 
Moreover, BIAS-ID enables a realistic comparison of different detectors and serves as a stress test for generalization and practical deployment.  
We validate our methodology by analyzing the presence of six biases across six state-of-the-art \ac{AIGI} detectors, utilizing two diverse datasets.
Our findings confirm that seemingly harmless transformations can drastically affect detection scores, emphasizing the need for bias-aware evaluation.

In summary, we make the following contributions:
\begin{enumerate}
    \item We address the critical but underexplored issue of bias in \ac{AIGI} detectors, focusing on biases induced by common image transformations.
    \item We develop a simple yet effective methodology to assess the presence of such biases. Unlike prior work, we derive sensible metrics directly from the predicted scores, allowing for a fine-grained analysis and more nuanced findings.
    \item Using this framework, we perform an exemplary analysis of six recently published \ac{AIGI} detectors, studying the presence of six biases across two datasets.
\end{enumerate}

\section{Related Work}
\label{sec:related}

\paragraph{Bias in AI-Generated Image Detectors}
The implications of training detectors on differently processed real and fake images have already been noted in early work on \ac{AIGI} detection~\cite{xuanGeneralizationGANImage2019,chaiWhatMakesFake2020}.
More recently, however, the issue of biased training data has received increased attention.
\citet{grommeltFakeJPEGRevealing2024} analyze how JPEG compression and image size can affect \ac{AIGI} detection.
They find that, because of differences in format and size between ImageNet~\cite{deng2009imagenet} and GenImage~\cite{zhuGenImageMillionscaleBenchmark2023}, detectors learn to distinguish real and fake images based on these properties rather than on actual forensic artifacts.
Their findings further show that removing these biases (e.g., by compressing generated images as well) can improve the detector's robustness and generalization.
Similar observations were made by \citet{choiCombatingDatasetMisalignment2025} regarding the popular ForenSynths~\cite{wangCNNgeneratedImagesAre2020} dataset.
\citet{lorchLandscapeMoreSecure2024} demonstrate that image orientation (landscape or portrait) can strongly affect the performance of forensic applications, including \ac{AIGI} detection.
\citet{rajanAlignedDatasetsImprove2025} show that if real and generated images differ in their resolutions, resizing them to the detector's input resolution can cause a bias w.r.t.\ up- and downscaling artifacts.

\paragraph{Mitigating Bias}
Data augmentation has traditionally been used to mitigate biases caused by imbalanced training data~\cite{wangCNNgeneratedImagesAre2020,xuanGeneralizationGANImage2019,chaiWhatMakesFake2020}.
However, the focus has recently shifted toward proactively avoiding dataset bias by aligning real and fake training samples.
\citet{cazenavetteFakeInversionLearningDetect2024} propose a novel data collection method to ensure semantic alignment.
Starting from a diverse set of generated images, they collect matching real images using reverse image search.
Other works approach the problem from the opposite direction: Given a dataset of real images, they create corresponding fake images.
A simple approach is to extract image descriptions from real images and use these to generate matching fake images~\cite{bammeySynthbusterDetectionDiffusion2023}.
Greater semantic alignment is achieved by leveraging the reconstruction capabilities of diffusion models to create fake images that are semantically near-identical to their real counterparts~\cite{chenDRCTDiffusionReconstruction2024,guillaroBiasfreeTrainingParadigm2025}.
A more efficient variant uses only the autoencoder of an LDM~\cite{rombachHighresolutionImageSynthesis2022} to generate semantically aligned fake images~\cite{rajanAlignedDatasetsImprove2025}.
Taken together, the goal is to transfer any semantic and processing-related properties from the real images to the fake images, encouraging the detector to rely on meaningful features rather than spurious correlations.
Detectors trained on such aligned datasets show improved generalization to unseen datasets~\cite{chenDRCTDiffusionReconstruction2024,rajanAlignedDatasetsImprove2025,guillaroBiasfreeTrainingParadigm2025}.

\paragraph{Robustness Evaluation}
The impact of image transformations is often discussed in terms of a detector's robustness. 
However, as we explain in the following section, robustness is a related but distinct concept.
It is usually assessed by applying the same transformation to real and fake images and measuring its effect on detection performance~\cite{wangCNNgeneratedImagesAre2020,cocchiUnveilingImpactImage2023,sabelRobustnessGeneralizabilityFace2021,gragnanielloAreGANGenerated2021,10.1145/3690624.3709392}.
Yet, as we show in our analysis, performance alone cannot always fully capture how a detector is affected by a transformation.

\section{What \emph{Is} a Bias in AI-Generated Image Detection?}
\label{sec:biases}

Dataset bias is a long-standing challenge in computer vision~\cite{TorralbaUnbiasedLook2011,FABBRIZZI2022103552}.
Training deep neural networks on biased datasets can lead to shortcuts, which are 
defined by~\citet{geirhosShortcutLearningDeep2020} as ``decision rules that perform well on i.i.d.\ test data but fail on o.o.d.\ tests, revealing a mismatch between intended and learned solution''.
In classical computer vision tasks, such as object recognition, this is typically related to the co-occurrence of different semantic concepts.
For instance, if cows frequently appear on grassland in a training set, a model trained on this biased dataset may incorrectly associate cows with grass, and consequently fail when presented with a cow in a different context (like a beach)~\cite{DBLP:conf/eccv/BeeryHP18}.

However, a well-trained object recognition model will correctly classify a cow regardless of whether the image has been compressed, resized, or otherwise processed, as long as the image's \emph{content} is not significantly altered.
In contrast, as generated images are often visually indistinguishable from real ones, an \ac{AIGI} detector should focus on \emph{low-level, imperceptible artifacts} introduced by image acquisition (real images) or generation (fake images).
This makes \ac{AIGI} detectors sensitive to image transformations. 
Processing operations like compression or resizing can also leave low-level traces or modify existing ones.
These artifacts often go unnoticed by humans and can also be challenging for automated methods to detect.
As a result, the training data may exhibit a systematic \emph{transformation bias}: Low-level image properties are distributed unevenly between real and generated images.
Similar to the cow-on-grass example, the detector may incorrectly associate a class label (real or fake) with the presence or absence of particular artifacts.

It is important to distinguish such biases from the more frequently discussed notion of \emph{lack of robustness}. 
A model is not robust to a transformation if the detection performance decreases, either due to a test-time distribution shift or the destruction of forensic traces.
Robustness can be partly improved through data augmentation, but very strong degradations can make reliable \ac{AIGI} classification simply infeasible.
Bias, in contrast, is caused by an unintended, subtle disparity between real and fake images during training, and can usually be avoided by aligning both datasets.

We finally note that the notion of semantic or content bias has also been discussed in \ac{AIGI} detection~\cite{cazenavetteFakeInversionLearningDetect2024,rajanAlignedDatasetsImprove2025,gyeReducingContentBias2025}. 
With modern text-to-image models trained on Internet-scale data, aligning the semantic distributions of real and fake images has become challenging.
Still, prior work has proposed various approaches for semantic alignment (see \Cref{sec:related}). Biases rooted in image transformations, on the other hand, are more subtle and harder to fully mitigate, which is why we focus on transformation biases in the remainder of this work.

\section{The BIAS-ID Analysis Framework}
\label{sec:methodology}

In this section, we present our methodology for analyzing biases in \ac{AIGI} detectors.
Unlike prior work, which typically infers the effects of bias and detector robustness from changes in classification performance, we argue that directly examining a classifier's predicted scores provides far richer insights.
We first introduce the concept of \emph{score shift} as a measure for assessing the susceptibility of a detector w.r.t.\ a transformation.
Afterward, we explain how to aggregate score shifts into a single-valued \emph{aggregated transformation sensitivity (ATS)} score.

\paragraph{Score Shift}
Given an image $x$, let $C(x) \in [0, 1]$ denote the score that a binary classifier $C$ assigns to it.
This score represents the predicted probability that the image is AI-generated: Lower scores indicate that an image is real, whereas higher scores indicate that it is AI-generated.
Applying a transformation $T$ (e.g., JPEG compression) at a certain level $l$ (e.g., quality) to $x$, we obtain $T_l(x)$.
Analogously, the classifier's prediction on the transformed image is denoted by $C(T_l(x))$.
We define the \emph{score shift} $\Delta_{T_l}$ as the difference between the classifier's score assigned to the transformed image and its score assigned to the original image:
\begin{equation}
    \Delta_{T_l}(x) = C(T_l(x)) - C(x) \enspace \in [-1, 1] \enspace .
\end{equation}
The score shift measures not only \emph{the magnitude} of a transformation's effect on the classifier, but also its \emph{direction}. 
If a transformation leads to a higher score, the score shift will be positive, indicating a tendency toward the fake class.
Conversely, a negative value for $\Delta_{T_l}(x)$ means that the transformation shifts the predictions toward the real class.
Note that, compared to using only the score assigned to the transformed image, analyzing score \emph{shifts} has the advantage that effects shared by the clean and transformed images, such as those caused by a distribution shift, cancel out.

To characterize the overall effect of a transformation across images, we additionally define the \emph{mean score shift (MSS)} $\overline{\Delta}_{T_l}$ over a set of images $X$. %
\begin{equation} 
\overline{\Delta}_{T_l}(X) = \frac{1}{|X|} \sum_{x \in X} \Delta_{T_l}(x) \enspace .
\end{equation}
Comparing the MSSs of a set of real images $X_R$ and a set of fake images $X_F$ provides detailed insights into how the classifier $C$ responds to a transformation $T_l$.
If $C$ is \emph{unbiased} w.r.t.~$T_l$, two cases can occur: 
First, $T_l$ may not significantly change the scores of both real and fake images, resulting in MSSs close to zero.
We can therefore conclude that $C$ is both \emph{robust} and \emph{unbiased} w.r.t.~$T_l$.
Second, the transformation may destroy or alter forensic traces used for detection, causing the scores for both real and fake images to shift toward $0.5$ by a similar amount.
In this case, $\overline{\Delta}_{T_l}(X_R)$ and $\overline{\Delta}_{T_l}(X_F)$ will have similar magnitudes but with opposite signs (i.e., $\overline{\Delta}_{T_l}(X_R) + \overline{\Delta}_{T_l}(X_F) \approx 0$).
This means that, while not being entirely robust w.r.t.~$T_l$ (because classification performance decreases), the detector $C$ does not suffer from a systematic bias toward a single class.

In contrast, if the MSSs for real and fake images are not close to zero and also do not cancel each other out (i.e., $\overline{\Delta}_{T_l}(X_R) + \overline{\Delta}_{T_l}(X_F) \neq 0$), we can conclude that $C$ is biased w.r.t.~$T_l$.
This means that, on average, applying the transformation $T_l$ to an image causes a systematic shift toward either the real or the fake class, distorting the classifier's predictions.

\paragraph{Aggregated Transformation Sensitivity}
Calculating the (mean) score shifts for real and fake images provides rich insights into a detector's behavior regarding a transformation $T_l$.
To facilitate the comparison between different detectors and datasets, we additionally introduce the \emph{aggregated transformation sensitivity (ATS)}, which summarizes the impact of a transformation $T$ over a given set of levels $L$ into a single value.
Given a set of images $X$, we compute $\text{ATS}_T$ by averaging the MSS across all levels $l \in L$:
\begin{equation}
 \text{ATS}_T(X) = \frac{1}{|L|} \sum_{l \in L} \overline{\Delta}_{T_l}(X) \enspace ,
\end{equation}
Analyzing this metric for both real images $X_R$ and fake images $X_F$ reveals the classifier's per-class shift tendency w.r.t.\ the transformation.
To obtain a detector's \emph{overall} sensitivity w.r.t.~$T$, taking into account both real and fake images, we additionally define
\begin{equation}
 \text{ATS}_T(X_R, X_F) = \frac{1}{|L|} \sum_{l \in L} \left|\overline{\Delta}_{T_l}(X_R) + \overline{\Delta}_{T_l}(X_F)\right| \enspace .
\end{equation}
Note that we take the absolute value of $\overline{\Delta}_{T_l}(X_R) + \overline\Delta_{T_l}(X_F)$ to account for the possibility that $C$'s predictions are shifted into opposite directions at different levels $l$.
Therefore, $\text{ATS}_T(X_R, X_F)$ ranges from $0$ to $2$, with higher values indicating a stronger sensitivity w.r.t.~$T$, regardless of the direction of the score shift.

\paragraph{Illustrative Example} 
We demonstrate the issue of transformation bias and how our methodology can identify it through a simple experiment.
We train a CLIP-based classifier~\cite{ojhaUniversalFakeImage2023} on 800 real (resized RAISE-1k~\cite{dang-nguyenRAISERawImages2015}) and 800 matching fake (generated with FLUX) images, both taken from~\cite{guillaroBiasfreeTrainingParadigm2025}.
200 images per class are held out as a test set.
We use JPEG compression at quality $80$ as an example for a potential transformation bias.
To study the effects of biased training data, we compare three settings.
Detector (a) uses JPEG augmentation during training, with quality randomly sampled from $[1,100]$.
Notably, augmentation is applied to both real and fake images.
In contrast, detector (b) is trained on compressed real (with fixed quality of $80$) and uncompressed fake images.
Conversely, detector (c) is trained on uncompressed real and compressed fake images.
In both cases, we do not apply any additional JPEG augmentation. 
We provide further training details in Appendix~\ref{app:experimental_details}.

\begin{wraptable}[6]{r}{132pt}
  \vspace{-0.75\baselineskip}
  \caption{\textbf{Per-class and overall ATS of the three detectors.}}
  \label{tab:illustrative_example_ats}
  \scriptsize
  \setlength{\tabcolsep}{\smalltabcolsep}
  \begin{tabular}{@{\ }c@{\hspace{2pt}}rrr@{\ }}
    \toprule
     & $\text{ATS}(X_R)$ & $\text{ATS}(X_F)$ & $\text{ATS}(X_R, X_F)$ \\
    \midrule
    (a) & 0.077 & -0.030 & 0.053  \\
    (b) & -0.489 & -0.571 & 1.059 \\
    (c) & 0.624 & 0.252 & 0.876 \\
    \bottomrule
  \end{tabular}
  \vspace{-3\baselineskip}
\end{wraptable}

\begin{figure*}[tbp]
    \centering
    \include{assets/figures/illustrative_example}
    \vspace{-22.5pt}
    \caption{\textbf{Demonstration of how training detectors on differently processed real and fake images can cause biased predictions.} For each detector (a)-(c), the plot on the left shows the mean scores and mean score shifts (MSSs) for real (\includegraphics[height=5pt]{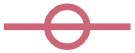}) and fake (\includegraphics[height=5pt]{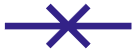}) images, with 95\% CIs depicted as shaded bands (very narrow, best viewed when zoomed in). The heatmap on the right depicts the detector's accuracy (assuming a decision threshold of $0.5$) when tested on different combinations of real and fake quality levels.}
    \label{fig:jpeg_validation}
\end{figure*}

The line plots in \Cref{fig:jpeg_validation} depict the mean scores and mean score shifts for all three detectors across different JPEG quality levels.
For detector (a), whose training data does not contain any intentional processing imbalance, scores remain relatively stable, resulting in MSSs close to zero.
While we observe a more pronounced effect at a quality of $10$, real and fake scores are shifted in opposite directions.
Thus, the detector is not systematically biased, but simply struggles when strong compression obscures the forensic artifacts.
This interpretation is reflected in the low ATS scores in \Cref{tab:illustrative_example_ats}.
In contrast, the detectors trained on biased data show clear signs of JPEG-related bias: With stronger JPEG compression, the predictions of detector (b) are shifted toward the real class, while the predictions of detector (c) are shifted toward the fake class.
Accordingly, we obtain negative/positive per-class ATS for detectors (b) and (c), respectively, and higher overall ATS than detector (a).
The consequences of using such biased detectors become apparent when looking at the accuracy heatmaps.
While the accuracy of detector (a) remains high for all combinations of real and fake JPEG quality levels, it strongly varies for detectors (b) and (c), making them unreliable under realistic conditions.

Our experiment also demonstrates why solely analyzing performance degradation---which is commonly used to evaluate a detector's robustness---is not sufficient to study the presence of biases.
A typical robustness evaluation would correspond to the diagonal (from bottom left to top right) of the heatmaps in \Cref{fig:jpeg_validation}.
While such an analysis would confirm that detector (a) is more robust, the results would be very similar for detectors (b) and (c), despite their distinct behavior.
Moreover, without inspecting scores and score shifts directly, one cannot distinguish between a lack of robustness (across both real and fake images) and a systematic bias toward a particular class.

\paragraph{Implementation}
We implement BIAS-ID in PyTorch~\cite{Ansel_PyTorch_2_Faster_2024} to ensure high compatibility with existing \ac{AIGI} detector implementations.
It is designed as an easy-to-use and extensible library that allows users to test their own detectors and add custom transformations and datasets.
We plan to continuously update and improve BIAS-ID to facilitate the bias-aware evaluation of future detection methods.
We therefore invite the scientific community to make suggestions or share their observations related to biases in \ac{AIGI} detectors.

\section{Evaluation}
\label{sec:evaluation}

In this section, we empirically validate our BIAS-ID framework through an evaluation of six recently proposed \ac{AIGI} detectors.
For each one, we quantify the presence of six typical transformation biases across two diverse datasets.

\subsection{Experimental Setup}

\paragraph{Data}
Choosing appropriate datasets for evaluating the presence of biases is challenging, since datasets may themselves contain unintended biases beyond the one under investigation, potentially confounding the results.
To mitigate such effects as much as possible, we select datasets that provide (i) high-quality images, (ii) semantic alignment between real and generated images, and (iii) a diverse range of scenes.
We primarily conduct our evaluation on real images from RAISE-1k~\cite{dang-nguyenRAISERawImages2015} and generated images from an extended version of Synthbuster~\cite{bammeySynthbusterDetectionDiffusion2023,guillaroBiasfreeTrainingParadigm2025}.
RAISE-1k is especially suitable because it comprises 1\,000 raw, uncompressed photos covering different scenes and objects.
Synthbuster~\cite{bammeySynthbusterDetectionDiffusion2023} is directly derived from RAISE-1k.
For each generator (DALL-E~2/3, Firefly\footnote{We exclude the Firefly subset in our evaluation, since, despite being saved as PNGs, they were found to be JPEGs, see \url{https://github.com/qbammey/synthbuster/issues/1}.}, Midjourney, Stable Diffusion~1.3/1.4/2/XL, GLIDE), images were created using prompts extracted from the corresponding real images with CLIP Interrogator~\cite{pharmapsychoticClipinterrogator2023} and Midjourney's ``describe'' feature.
We additionally include the extension by \citet{guillaroBiasfreeTrainingParadigm2025}, who generated images with FLUX and Stable Diffusion~3.5 following the same methodology.
To align the resolution of real and fake images, we use the resized version of RAISE-1k~\cite{guillaroBiasfreeTrainingParadigm2025}.
In Appendix~\ref{app:raise1k_vs_resizing} we analyze the effect of resizing by performing additional experiments with original-size images.

To assess the consistency of our findings, we use CLIC2020~\cite{clic2020} and SynthCLIC~\cite{williSyntheticImageDetection2026} as an additional dataset pair.
CLIC2020 contains 2\,163 real images from both professional cameras and mobile phones.
Consequently, the real images may already have undergone compression, resizing, or other types of post-processing.
For each real image, fake images were generated with four models (Imagen~3, FluxDev, FluxSchnell, Stable Diffusion~3 Medium) using prompts extracted with Gemini (gemini-2.0-flash-001).

\paragraph{Detectors}

\begin{wraptable}[9]{r}{168pt}
  \vspace{-1.85\baselineskip}
  \caption{\textbf{Detectors included in our bias evaluation.} Note that DRCT was trained on a variety of different diffusion models.}
  \label{tab:detectors}
  \scriptsize
  \begin{tabular}{@{\ }lll@{\ }}
    \toprule
    Detector & Venue & Real/Fake Training Data \\
    \midrule
    UnivFD~\cite{ojhaUniversalFakeImage2023} & CVPR 2023 & LSUN/ProGAN\\
    DRCT~\cite{chenDRCTDiffusionReconstruction2024} & ICML 2024 & COCO/DMs \\
    RINE~\cite{koutlisLeveragingRepresentationsIntermediate2025} & ECCV 2024 & LSUN/ProGAN \\
    AIDE~\cite{yanSanityCheckAIgenerated2024a} & ICLR 2025 & ImageNet/SD1.4 \\
    SPAI~\cite{karageorgiouAnyResolutionAIGeneratedImage2025} & CVPR 2025 & LSUN/LDM \\
    B-Free~\cite{guillaroBiasfreeTrainingParadigm2025} & CVPR 2025 & COCO/SD2.1 \\
    \bottomrule
  \end{tabular}
\end{wraptable}

\Cref{tab:detectors} lists the six detectors we include in our evaluation.
UnivFD~\cite{ojhaUniversalFakeImage2023} uses a single linear layer on top of features extracted by CLIP~\cite{radfordLearningTransferableVisual2021} to avoid overfitting on a single generative model, which generalizes better than fully trained classifiers.
DRCT~\cite{chenDRCTDiffusionReconstruction2024} also uses CLIP as a feature extractor, but was trained on real and ``diffusion-reconstructed'' images to ensure semantic alignment within the training data.
RINE~\cite{koutlisLeveragingRepresentationsIntermediate2025} uses features from intermediate CLIP layers (instead of only the final embedding) to capture low-level image features.
AIDE~\cite{yanSanityCheckAIgenerated2024a} combines high-level CLIP features with low-level artifacts from high- and low-frequency patches.
In contrast, SPAI~\cite{karageorgiouAnyResolutionAIGeneratedImage2025} focuses on frequency artifacts: By training a spectral reconstruction model on real images, \acp{AIGI} can be detected due to their higher reconstruction error.
Lastly, B-Free~\cite{guillaroBiasfreeTrainingParadigm2025} proposes a bias-free training paradigm.
Besides ensuring semantic alignment by conditioning the generator on real images (similar to DRCT), it uses content-based augmentation to further improve generalization.
For each detector, we use code and checkpoints provided by the respective authors.
Importantly, we use the original input processing (e.g., cropping, resizing, normalization) after applying each test-time transformation.

\paragraph{Transformations}
We consider six common image transformations that are likely to occur in training data or during real-world deployment of detectors. %
Due to its widespread use on the Internet, JPEG compression is one of the most important transformations.
In our evaluation, we additionally include WebP as an alternative compression technique.
For both formats, we analyze quality levels from $100$ (high quality) to $10$ (low quality).
Another highly relevant transformation is resizing.
We analyze up- and downscaling separately to capture distinct trends within our ATS scores.
We consider resizing factors from $0.5$ to $2.0$ (with $1.0$ referring to the original size), and apply both bilinear and bicubic resizing (the latter in Appendix~\ref{app:bicubic_resizing}) to study the effect of the resizing algorithm.
While not as subtle as compression and resizing, we include rotation as another common image transformation.
Lastly, we analyze how converting images to grayscale affects the detector's predictions.

\paragraph{Computational Resources}
Preliminary runs and the experiments shown in \Cref{fig:jpeg_validation} were conducted on a single NVIDIA A40 GPU (48\,GB).
The main evaluation was performed on an HPC cluster, with each job being allocated one NVIDIA A30 GPU (24\,GB), eight CPU cores and 62.5\,GB of memory.
Since our framework only performs inference, all experiments are reproducible with less powerful hardware, as long as the computational requirements of the detector under test are met.
The total runtime of all experiments reported in this paper was approximately two weeks.

\subsection{Results}
For each detector, transformation, and dataset pair, we evaluate bias sensitivity using our methodology introduced in \Cref{sec:methodology}.
We analyze fake images from each generator separately and report the mean over all subsets.
To compare our proposed ATS score with traditional performance degradation measures, we additionally compute the mean accuracy/AUC shift across all levels (compared to performance on clean images).
We use 10\,000 bootstrap resamples to compute confidence intervals.

\paragraph{Compression}
We first analyze JPEG compression in \Cref{fig:synthbuster_jpeg}.
B-Free, which employs different strategies to avoid dataset bias, indeed produces stable predictions at low and moderate compression levels.
However, we observe that sensitivity is higher when tested on CLIC2020\,/\,SynthCLIC (see \Cref{fig:synthclic_jpeg} in the appendix), especially for fake images.
This is also reflected in the higher magnitude of $\text{ATS}(X_F)$ and subsequently $\text{ATS}(X_R,X_F)$.
UnivFD, DRCT, and RINE are only moderately biased, yet their performance fluctuates and decreases under stronger compression.
Especially the results for UnivFD show that a detector can be unbiased but still ineffective at the same time.
The analysis on the second dataset pair suggests that real and fake images are harder to distinguish; nevertheless, ATS scores are largely consistent.
For AIDE and SPAI, we observe a strong correlation between JPEG compression and lower scores, which is further supported by the negative per-class ATS scores.
A possible explanation is that both methods rely on frequency information, which is affected by JPEG (in particular high-frequency components, which often contain forensic traces~\cite{corviIntriguingPropertiesSynthetic2023}).

\begin{figure}[htbp]
    \centering
    \include{assets/figures/tex/raise1k_resized_vs_synthbuster/jpeg}
    \vspace{-10pt}
    \caption{\textbf{Analysis of JPEG bias.} The top row shows mean score, mean score shift (MSS), and AUC, all with 95\% CIs as shaded bands (very narrow, best viewed when zoomed in) for \datasetcaption{RAISE-1k}{Synthbuster}. We provide plots for CLIC2020\,/\,SynthCLIC in Appendix~\ref{app:synthclic} and exact numerical values in Appendix~\ref{app:plot_values}. The bottom table shows ATS (with 95\% CIs) and mean accuracy/AUC shift (denoted as $\Delta\text{Acc.}$ and $\Delta\text{AUC}$).}
    \label{fig:synthbuster_jpeg}
\end{figure}

\begin{figure}[htbp]
    \centering
    \include{assets/figures/tex/raise1k_resized_vs_synthbuster/webp}
    \vspace{-10pt}
    \caption{\textbf{Analysis of WebP bias.} The plots show results for \datasetcaption{RAISE-1k}{Synthbuster}.}
    \label{fig:synthbuster_webp}
\end{figure}

\Cref{fig:synthbuster_webp} contains the results for WebP compression.
DRCT, RINE, and SPAI are notably more affected by WebP than by JPEG, with DRCT showing a positive correlation between compression level and score.
We hypothesize that detectors have not encountered WebP-compressed images during training, which leads to different out-of-distribution behavior.
For both compression algorithms, AIDE serves as a good example of why performance degradation alone is not sufficient for bias analysis.
While scores and score shifts clearly indicate a bias toward the real class, $\Delta \text{Acc.}$ and $\Delta \text{AUC}$ are both positive (for RAISE-1k\,/\,Synthbuster), suggesting better classification with increasing compression strength.
However, in a realistic setting, the prediction of such a biased detector predominantly depends on the compression level, not whether the image is actually real or generated.

\paragraph{Resizing}
\Cref{fig:synthbuster_upscaling_bilinear} shows that UnivFD, DRCT, and SPAI tend to assign higher scores to upscaled images.
For B-Free, we observe this trend only for SynthCLIC (see \Cref{fig:synthclic_upscaling_bilinear}).
Interestingly, the predictions of AIDE fluctuate strongly, resulting in large overall ATS scores.
While it is difficult to explain this behavior, a possible reason might be that AIDE does not include resizing augmentation during training.
When analyzing downscaling (see \Cref{fig:synthbuster_downscaling_bilinear}), UnivFD and SPAI behave consistently with the analysis of upscaling, i.e., scores decrease as the resizing factor becomes smaller.
Downscaling also appears to have a stronger effect on B-Free across both dataset pairs.
Compared with upscaling, AIDE produces more stable scores, similar to DRCT and RINE.
Our additional analysis of bicubic (instead of bilinear) up- and downscaling (see Appendix~\ref{app:bicubic_resizing}) suggests that the choice of resizing algorithm does not affect the qualitative trends observed.
However, it appears that bicubic resizing results in ATS scores with higher magnitudes.

\begin{figure}[htbp]
    \centering
    \include{assets/figures/tex/raise1k_resized_vs_synthbuster/upscaling_bilinear}
    \vspace{-10pt}
    \caption{\textbf{Analysis of bilinear upscaling bias.} The plots show results for \datasetcaption{RAISE-1k}{Synthbuster}.}
    \label{fig:synthbuster_upscaling_bilinear}
\end{figure}

\begin{figure}[htbp]
    \centering
    \include{assets/figures/tex/raise1k_resized_vs_synthbuster/downscaling_bilinear}
    \vspace{-10pt}
    \caption{\textbf{Analysis of bilinear downscaling bias.} The plots show results for \datasetcaption{RAISE-1k}{Synthbuster}.}
    \label{fig:synthbuster_downscaling_bilinear}
\end{figure}

\newpage
\paragraph{Rotation}
\Cref{fig:synthbuster_rotation} shows that all detectors are affected by rotation, with DRCT, SPAI, and B-Free exhibiting a repeating pattern.
Our results for both dataset pairs are largely consistent, with lower ATS scores for CLIC2020\,/\,SynthCLIC.
The high sensitivity to rotation may again be related to the fact that rotation is not commonly used as a training-time augmentation.
Our findings are consistent with those of \citet{lorchLandscapeMoreSecure2024}, corroborating the importance of image orientation for forensic applications.
However, while their work attributes the drop in detection performance to detectors overfitting on directional statistics of generated images (caused by the autoencoder), our analysis of raw scores shows that real images are even more affected, with an average ATS of $0.162$ for real images (across both dataset pairs), compared with $0.13$ for fake images.
While the overall qualitative trends are similar for both dataset pairs, it appears that real images from CLIC2020 are less affected (see \Cref{fig:synthclic_rotation} in the appendix).

\begin{figure}[htbp]
    \centering
    \include{assets/figures/tex/raise1k_resized_vs_synthbuster/rotation}
    \vspace{-10pt}
    \caption{\textbf{Analysis of rotation bias.} The plots show results for \datasetcaption{RAISE-1k}{Synthbuster}.}
    \label{fig:synthbuster_rotation}
\end{figure}

\paragraph{Grayscale Conversion}
Finally, we analyze the detectors' sensitivity w.r.t.\ an image's color space in \Cref{fig:synthbuster_grayscale}.
Most detectors are not significantly biased, although detection performance noticeably declines for all except B-Free.
For DRCT, we observe a slight tendency to wrongly predict grayscale real images as fake.
In contrast, RINE and SPAI are more severely affected and tend to misclassify grayscale fake images as real.
Because RINE uses intermediate CLIP representations, it may be more sensitive to color information than detectors that use only the final-layer embedding.
For SPAI, one possible explanation is that the learned spectral model of real images associates the absence of color with the real class due to grayscale images in the training data.
We observe consistent qualitative trends for CLIC2020\,/\,SynthCLIC (see \Cref{fig:synthclic_grayscale} in the appendix).

\begin{figure}[htbp]
    \centering
    \include{assets/figures/tex/raise1k_resized_vs_synthbuster/grayscale}
    \vspace{-10pt}
    \caption{\textbf{Analysis of grayscale conversion bias.} The plots show results for \datasetcaption{RAISE-1k}{Synthbuster}.}
    \label{fig:synthbuster_grayscale}
\end{figure}

\newpage
\section{Discussion}
\label{sec:discussion}
\paragraph{Key Takeaways}
Our evaluation shows that transformation biases in \ac{AIGI} detectors are a common phenomenon that can affect their performance when tested on real-world data.
We observe that common transformations, such as compression, resizing, rotation, and grayscale conversion can cause systematic score shifts toward the real or the fake class.
Although we can confirm that some detectors are relatively insensitive to individual transformations (e.g., B-Free to JPEG), none of them is completely unbiased.
Importantly, the effects of bias are not always visible from inspecting performance metrics alone.
As we demonstrate, AUC curves or mean performance degradation are often not sufficient to understand a detector's behavior.
By relying on score shifts instead, our framework is able to analyze whether a detector is just less certain under a transformation, or whether it has a systematic bias that consistently favors one class.
This distinction can help researchers develop countermeasures, such as adjusting the detector's training dataset or optimization procedure.

\paragraph{Limitations}
Our work has several limitations. 
Although we evaluate six transformations across six state-of-the-art \ac{AIGI} detectors and two diverse dataset pairs, our analysis is not exhaustive.
In particular, \ac{AIGI} detectors might contain additional biases we did not consider.
Moreover, despite taking great care to ensure a controlled analysis, we cannot rule out that the tested datasets contain unknown biases, and other datasets would yield different results.
While our comparison between the two dataset pairs shows that the overall trends are generally consistent, there are deviations for some detector/bias combinations.
Nevertheless, the narrow 95\% CIs support the reliability of most reported findings.
Our exemplary sample size analysis in Appendix~\ref{app:sample_size_analysis} additionally suggests that one can obtain reliable estimates with even fewer samples.
In future work, we plan to extend our analysis and include images from up-to-date generators.
Due to the continuous development of generative models, it is unfortunately challenging to establish a single benchmark dataset.
Lastly, our methodology relies on comparing the detection score between an original and a transformed image.
As a consequence, our framework is not applicable to content biases (see \Cref{sec:biases}), since the property is inherent to the image itself and cannot be easily modified.
We consider the development of alternative metrics that can quantify semantic biases an important direction for future work.

\paragraph{Societal Impact}
We envision BIAS-ID as a useful tool for researchers to develop more robust and reliable \ac{AIGI} detectors, ultimately mitigating the harmful implications of generative AI.
However, adversaries may misuse our framework to find weaknesses in existing, open-source detectors, allowing them to bypass authenticity checks.
Given the potential benefits for improving future detection methods, we deem this risk acceptable.

\section{Conclusion}
\label{sec:conclusion}

We present BIAS-ID, a comprehensive framework for identifying and measuring transformation biases in \ac{AIGI} detectors.
Instead of inspecting performance degradation alone, we propose a score-based analysis.
By introducing mean score shift and aggregated transformation sensitivity (ATS), our method enables us to quantify whether and to what extent a detector is biased.
Applying BIAS-ID to six detectors across two datasets, we find that several detectors have systematic biases toward the real or the fake class under particular image transformations. 
These findings underline that bias-aware evaluation should become a routine part of \ac{AIGI} detection research.
We see BIAS-ID as a first step in this direction, enabling researchers to assess more precisely whether their detector is susceptible to transformation biases.
The framework is modular and can be extended with further transformations, datasets, and detectors.

\begin{ack}
We thank Luisa Verdoliva, Davide Cozzolino, and Fabrizio Guillaro for the discussions and their helpful suggestions. 
This work is funded by the Deutsche Forschungsgemeinschaft (DFG, German Research Foundation) under Germany's Excellence Strategy - EXC 2092 CASA - 390781972 and by the Ministry of Culture and Science of North Rhine-Westphalia as part of the Lamarr Fellow Network.
Calculations for this publication were partly performed on the HPC cluster Elysium of the Ruhr University Bochum, subsidized by the DFG (INST 213/1055-1).
\end{ack}

\medskip

\bibliographystyle{plainnat}
\bibliography{macros/conferences, macros/journals,main}

\appendix

\section{Experimental Details}
\label{app:experimental_details}

\paragraph{Training Procedure}
The CLIP-based detector used within the illustrative example (see \Cref{sec:methodology}) consists of a pre-trained, frozen CLIP-ViT-L/14 followed by a single fully-connected layer.
During training, we randomly crop images to the input resolution of $224\times224$ pixels.
During testing, we follow the standard approach to take the center crop.
The model is optimized for $10$ epochs using AdamW~\cite{loshchilovDecoupledWeightDecay2018} with a learning rate of $0.001$.

\paragraph{Dataset Licenses}
The RAISE dataset~\cite{dang-nguyenRAISERawImages2015} may be used for ``non-commercial research and educational purposes''\footnote{\url{https://loki.disi.unitn.it/RAISE/download.html}}.
Synthbuster~\cite{bammeySynthbusterDetectionDiffusion2023} is released under a CC BY-NC-SA 4.0 license\footnote{\url{https://zenodo.org/records/10066460}}.
The professional set of CLIC2020 is released under Unsplash's license, while the mobile set uses a CC0 1.0 license\footnote{\url{https://groups.google.com/g/clic-2020/c/pVCa3pH0nQk/m/UhPiILHVAQAJ}}.
SynthCLIC~\cite{williSyntheticImageDetection2026} is released under an unspecified license from the Creative Commons license family\footnote{\url{https://huggingface.co/datasets/marco-willi/synthclic}}.

\section{Sample Size Analysis}
\label{app:sample_size_analysis}
To analyze how many samples are required to obtain meaningful ATS scores, we conduct an exemplary sample size analysis for JPEG compression on resized RAISE-1k\,/\,Synthbuster.
\Cref{tab:sample_size_analysis} reports ATS, 95\% CI and 95\% CI width depending on the number of samples per generator.
The results for 1\,000 samples correspond to those in \Cref{fig:synthbuster_jpeg}.
While we observe greater deviations when using only 50 or 100 samples, values obtained from 500 samples are generally close to our final results.

\begin{table}[htbp]
    \centering
    \tiny
    \setlength{\tabcolsep}{\smalltabcolsep}
    \caption{\textbf{Analysis of how ATS scores and CIs are affected by varying numbers of samples per generator.} Computed for JPEG compression on resized RAISE-1k\,/\,Synthbuster.}
    \label{tab:sample_size_analysis}
    \begin{tabular}{@{\ }lr@{\quad}rrr@{\quad}rrr@{\quad}rrr@{\ }}
\toprule
& 
& \multicolumn{3}{c}{$\text{ATS}(X_R)$}
& \multicolumn{3}{c}{$\text{ATS}(X_F)$} & \multicolumn{3}{c}{$\text{ATS}(X_R, X_F)$} \\
\cmidrule(l{0pt}r{6pt}){3-5} \cmidrule(l{0pt}r{6pt}){6-8} \cmidrule(l{0pt}r{1pt}){9-11}
& \#Samples
& Estimate & 95\% CI & 95\% CI Width
& Estimate & 95\% CI & 95\% CI Width
& Estimate & 95\% CI & 95\% CI Width \\
\midrule
UnivFD & 50 & 0.071 & (0.051, 0.093) & 0.042 & -0.067 & (-0.085, -0.049) & 0.036 & 0.170 & (0.152, 0.192) & 0.040 \\
 & 100 & 0.073 & (0.051, 0.097) & 0.046 & -0.072 & (-0.086, -0.059) & 0.027 & 0.170 & (0.156, 0.186) & 0.029 \\
 & 500 & 0.063 & (0.054, 0.072) & 0.019 & -0.072 & (-0.078, -0.066) & 0.013 & 0.149 & (0.142, 0.155) & 0.013 \\
 & 1000 & 0.063 & (0.056, 0.070) & 0.014 & -0.073 & (-0.078, -0.068) & 0.009 & 0.151 & (0.146, 0.155) & 0.009 \\
\midrule
DRCT & 50 & 0.253 & (0.198, 0.306) & 0.108 & -0.061 & (-0.076, -0.046) & 0.030 & 0.216 & (0.175, 0.265) & 0.090 \\
 & 100 & 0.214 & (0.176, 0.254) & 0.079 & -0.060 & (-0.070, -0.049) & 0.021 & 0.193 & (0.168, 0.223) & 0.055 \\
 & 500 & 0.217 & (0.202, 0.232) & 0.030 & -0.065 & (-0.069, -0.061) & 0.009 & 0.185 & (0.175, 0.195) & 0.020 \\
 & 1000 & 0.218 & (0.207, 0.228) & 0.021 & -0.065 & (-0.068, -0.062) & 0.006 & 0.184 & (0.177, 0.191) & 0.015 \\
\midrule
RINE & 50 & 0.161 & (0.128, 0.195) & 0.066 & -0.146 & (-0.160, -0.131) & 0.030 & 0.153 & (0.146, 0.166) & 0.020 \\
 & 100 & 0.153 & (0.129, 0.176) & 0.046 & -0.145 & (-0.157, -0.134) & 0.023 & 0.145 & (0.139, 0.153) & 0.014 \\
 & 500 & 0.162 & (0.152, 0.172) & 0.019 & -0.149 & (-0.154, -0.144) & 0.010 & 0.147 & (0.144, 0.151) & 0.007 \\
 & 1000 & 0.149 & (0.141, 0.156) & 0.015 & -0.148 & (-0.152, -0.145) & 0.007 & 0.150 & (0.148, 0.152) & 0.005 \\
\midrule
AIDE & 50 & -0.617 & (-0.717, -0.514) & 0.203 & -0.512 & (-0.534, -0.488) & 0.046 & 1.128 & (1.014, 1.237) & 0.223 \\
 & 100 & -0.532 & (-0.605, -0.456) & 0.149 & -0.520 & (-0.535, -0.505) & 0.030 & 1.052 & (0.973, 1.128) & 0.155 \\
 & 500 & -0.530 & (-0.563, -0.496) & 0.067 & -0.527 & (-0.534, -0.520) & 0.014 & 1.057 & (1.022, 1.092) & 0.071 \\
 & 1000 & -0.547 & (-0.571, -0.523) & 0.048 & -0.528 & (-0.532, -0.523) & 0.009 & 1.074 & (1.050, 1.099) & 0.050 \\
\midrule
SPAI & 50 & 0.209 & (0.133, 0.289) & 0.156 & -0.202 & (-0.223, -0.182) & 0.041 & 0.311 & (0.269, 0.369) & 0.100 \\
 & 100 & 0.124 & (0.066, 0.183) & 0.117 & -0.226 & (-0.243, -0.209) & 0.034 & 0.291 & (0.271, 0.319) & 0.048 \\
 & 500 & 0.104 & (0.081, 0.128) & 0.048 & -0.225 & (-0.233, -0.216) & 0.017 & 0.272 & (0.263, 0.283) & 0.020 \\
 & 1000 & 0.099 & (0.083, 0.117) & 0.034 & -0.222 & (-0.228, -0.216) & 0.012 & 0.275 & (0.268, 0.283) & 0.015 \\
\midrule
B-Free & 50 & 0.056 & (0.039, 0.079) & 0.040 & -0.109 & (-0.121, -0.096) & 0.025 & 0.085 & (0.078, 0.097) & 0.019 \\
 & 100 & 0.057 & (0.043, 0.074) & 0.031 & -0.098 & (-0.107, -0.089) & 0.018 & 0.078 & (0.073, 0.087) & 0.014 \\
 & 500 & 0.045 & (0.040, 0.050) & 0.011 & -0.095 & (-0.098, -0.091) & 0.008 & 0.077 & (0.074, 0.081) & 0.007 \\
 & 1000 & 0.045 & (0.042, 0.049) & 0.008 & -0.093 & (-0.096, -0.091) & 0.006 & 0.075 & (0.073, 0.078) & 0.005 \\
\bottomrule
\end{tabular}

\end{table}

\newpage
\section{Plots for CLIC2020\,/\,SynthCLIC}
\label{app:synthclic}
We show each detector's mean score, mean score shift (MSS) and AUC for CLIC2020\,/\,SynthCLIC, analogous to \Cref{fig:synthbuster_jpeg,fig:synthbuster_webp,fig:synthbuster_upscaling_bilinear,fig:synthbuster_downscaling_bilinear,fig:synthbuster_rotation,fig:synthbuster_grayscale}.

\begin{figure}[htbp]
    \centering
    \include{assets/figures/tex/clic2020_all_vs_synthclic_all/jpeg}
    \vspace{-22.5pt}
    \caption{\textbf{Analysis of JPEG bias for \datasetcaption{CLIC2020}{SynthCLIC}.}}
    \label{fig:synthclic_jpeg}
\end{figure}

\begin{figure}[htbp]
    \centering
    \include{assets/figures/tex/clic2020_all_vs_synthclic_all/webp}
    \vspace{-22.5pt}
    \caption{\textbf{Analysis of WebP bias for \datasetcaption{CLIC2020}{SynthCLIC}.}}
    \label{fig:synthclic_webp}
\end{figure}

\begin{figure}[htbp]
    \centering
    \include{assets/figures/tex/clic2020_all_vs_synthclic_all/upscaling_bilinear}
    \vspace{-22.5pt}
    \caption{\textbf{Analysis of bilinear upscaling bias for \datasetcaption{CLIC2020}{SynthCLIC}.}}
    \label{fig:synthclic_upscaling_bilinear}
\end{figure}

\begin{figure}[h!]
    \centering
    \include{assets/figures/tex/clic2020_all_vs_synthclic_all/downscaling_bilinear}
    \vspace{-22.5pt}
    \caption{\textbf{Analysis of bilinear downscaling bias for \datasetcaption{CLIC2020}{SynthCLIC}.}}
    \label{fig:synthclic_downscaling_bilinear}
\end{figure}

\begin{figure}[h!]
    \centering
    \include{assets/figures/tex/clic2020_all_vs_synthclic_all/rotation}
    \vspace{-22.5pt}
    \caption{\textbf{Analysis of rotation bias for \datasetcaption{CLIC2020}{SynthCLIC}.}}
    \label{fig:synthclic_rotation}
\end{figure}

\begin{figure}[h!]
    \centering
    \include{assets/figures/tex/clic2020_all_vs_synthclic_all/grayscale}
    \vspace{-22.5pt}
    \caption{\textbf{Analysis of grayscale conversion bias for \datasetcaption{CLIC2020}{SynthCLIC}.}}
    \label{fig:synthclic_grayscale}
\end{figure}

\newpage
\section{Results for Bicubic Resizing}
\label{app:bicubic_resizing}
To analyze whether the choice of resizing algorithm affects our results, we repeat our evaluation using bicubic instead of bilinear resizing.
\Cref{fig:upscaling_bicubic,fig:downscaling_bicubic} show the results for both dataset pairs.

\begin{figure}[htbp]
    \centering
    \include{assets/figures/tex/upscaling_bicubic}
    \vspace{-10pt}
    \caption{\textbf{Analysis of bicubic upscaling bias.} The top row shows results for \datasetcaption{RAISE-1k}{Synthbuster}, while the middle row shows results for \datasetcaption{CLIC2020}{SynthCLIC}.}
    \label{fig:upscaling_bicubic}
\end{figure}

\begin{figure}[htbp]
    \centering
    \include{assets/figures/tex/downscaling_bicubic}
    \vspace{-10pt}
    \caption{\textbf{Analysis of bicubic downscaling bias.} The top row shows results for \datasetcaption{RAISE-1k}{Synthbuster}, while the middle row shows results for \datasetcaption{CLIC2020}{SynthCLIC}.}
    \label{fig:downscaling_bicubic}
\end{figure}

\newpage
\section{Comparison of Resized and Original RAISE-1k}
\label{app:raise1k_vs_resizing}
Due to the high resolution of the original RAISE-1k dataset, we use the resized version provided by \citet{guillaroBiasfreeTrainingParadigm2025} in our main evaluation.
We repeat the bias analyses for JPEG and bilinear resizing using the original, full-size images from RAISE-1k (see \Cref{fig:raise1k_vs_resized_jpeg,fig:raise1k_vs_resized_upscaling_bilinear,fig:raise1k_vs_resized_downscaling_bilinear}).
UnivFD, DRCT, and RINE assign higher scores to the original RAISE-1k images, leading to reduced detection performance.
Except for the effect of JPEG compression on AIDE, we observe no substantial differences.

\enlargethispage{\baselineskip}

\begin{figure}[htbp]
    \centering
    \include{assets/figures/tex/raise1k_vs_synthbuster/jpeg}
    \vspace{-10pt}
    \caption{\textbf{Analysis of JPEG compression bias.} The top row shows results for \datasetcaption{resized RAISE-1k}{Synthbuster}, while the middle row shows results for \datasetcaption{original RAISE-1k}{Synthbuster}.}
    \label{fig:raise1k_vs_resized_jpeg}
\end{figure}

\begin{figure}[!htbp]
    \centering
    \include{assets/figures/tex/raise1k_vs_synthbuster/upscaling_bilinear}
    \vspace{-10pt}
    \caption{\textbf{Analysis of bilinear upscaling bias.} The top row shows results for \datasetcaption{resized RAISE-1k}{Synthbuster}, while the middle row shows results for \datasetcaption{original RAISE-1k}{Synthbuster}.}
    \label{fig:raise1k_vs_resized_upscaling_bilinear}
\end{figure}

\begin{figure}[!htbp]
    \centering
    \include{assets/figures/tex/raise1k_vs_synthbuster/downscaling_bilinear}
    \vspace{-10pt}
    \caption{\textbf{Analysis of bilinear downscaling bias.} The top row shows results for \datasetcaption{resized RAISE-1k}{Synthbuster}, while the middle row shows results for \datasetcaption{original RAISE-1k}{Synthbuster}.}
    \label{fig:raise1k_vs_resized_downscaling_bilinear}
\end{figure}

\clearpage
\section{Numerical Values of Plotted Metrics}
\label{app:plot_values}

Below we provide the exact values for mean score, mean score shift (MSS), and AUC we used to generate our plots.
\Cref{tab:raise1k_resized_vs_synthbuster_plot_values} corresponds to \Cref{fig:synthbuster_jpeg,fig:synthbuster_webp,fig:synthbuster_upscaling_bilinear,fig:synthbuster_downscaling_bilinear,fig:synthbuster_rotation,fig:synthbuster_grayscale} and the top rows of \Cref{fig:upscaling_bicubic,fig:downscaling_bicubic}.
\Cref{tab:clic2020_all_vs_synthclic_all_plot_values} corresponds to \Cref{fig:synthclic_jpeg,fig:synthclic_webp,fig:synthclic_upscaling_bilinear,fig:synthclic_downscaling_bilinear,fig:synthclic_rotation,fig:synthclic_grayscale} and the middle rows of \Cref{fig:upscaling_bicubic,fig:downscaling_bicubic}.
\Cref{tab:raise1k_vs_synthbuster_plot_values} corresponds to \Cref{fig:raise1k_vs_resized_jpeg,fig:raise1k_vs_resized_upscaling_bilinear,fig:raise1k_vs_resized_downscaling_bilinear}.

\begin{table}[htbp]
    \centering
    \fontsize{5}{5}\selectfont
    \setlength{\tabcolsep}{1.05pt}
    \caption{\textbf{Numerical values of the plots for RAISE-1k (resized)\,/\,Synthbuster.}}
    \label{tab:raise1k_resized_vs_synthbuster_plot_values}
    \begin{tabular}{@{\ }p{30pt}rrrrrrrrrrrrrrrrrrrrrrrrrrrrrrr@{\ }}
\toprule
 & & \multicolumn{5}{c}{UnivFD} & \multicolumn{5}{c}{DRCT} & \multicolumn{5}{c}{RINE} & \multicolumn{5}{c}{AIDE} & \multicolumn{5}{c}{SPAI} & \multicolumn{5}{c}{B-Free} \\
 \cmidrule(l{1pt}r{1pt}){3-7} \cmidrule(l{1pt}r{1pt}){8-12} \cmidrule(l{1pt}r{1pt}){13-17} \cmidrule(l{1pt}r{1pt}){18-22} \cmidrule(l{1pt}r{1pt}){23-27} \cmidrule(l{1pt}r{1pt}){28-32}
 & & \multicolumn{2}{c}{Score} & \multicolumn{2}{c}{MSS} & AUC & \multicolumn{2}{c}{Score} & \multicolumn{2}{c}{MSS} & AUC & \multicolumn{2}{c}{Score} & \multicolumn{2}{c}{MSS} & AUC & \multicolumn{2}{c}{Score} & \multicolumn{2}{c}{MSS} & AUC & \multicolumn{2}{c}{Score} & \multicolumn{2}{c}{MSS} & AUC & \multicolumn{2}{c}{Score} & \multicolumn{2}{c}{MSS} & AUC \\
\cmidrule(r{1pt}){3-4} \cmidrule(l{1pt}){5-6}
\cmidrule(r{1pt}){8-9} \cmidrule(l{1pt}){10-11}
\cmidrule(r{1pt}){13-14} \cmidrule(l{1pt}){15-16}
\cmidrule(r{1pt}){18-19} \cmidrule(l{1pt}){20-21}
\cmidrule(r{1pt}){23-24} \cmidrule(l{1pt}){25-26}
\cmidrule(r{1pt}){28-29} \cmidrule(l{1pt}){30-31}
 Transform & Level & Real & Fake & Real & Fake &  & Real & Fake & Real & Fake &  & Real & Fake & Real & Fake &  & Real & Fake & Real & Fake &  & Real & Fake & Real & Fake &  & Real & Fake & Real & Fake &  \\
\midrule
\multirow[t]{11}{*}{JPEG} & - & .07 & .34 & .00 & .00 & .75 & .35 & .84 & .00 & .00 & .93 & .25 & .76 & .00 & .00 & .90 & .64 & .80 & .00 & .00 & .62 & .12 & .87 & .00 & .00 & .95 & .02 & .93 & .00 & .00 & 1.0 \\
 & 100 & .06 & .30 & -.01 & -.05 & .75 & .53 & .86 & .18 & .02 & .88 & .27 & .71 & .02 & -.05 & .89 & .14 & .50 & -.50 & -.30 & .78 & .15 & .87 & .03 & .01 & .94 & .02 & .94 & .00 & .01 & 1.0 \\
 & 90 & .05 & .22 & -.02 & -.12 & .72 & .41 & .68 & .07 & -.17 & .75 & .25 & .51 & .00 & -.25 & .79 & .07 & .14 & -.58 & -.66 & .62 & .31 & .88 & .19 & .01 & .89 & .03 & .93 & .01 & .00 & 1.0 \\
 & 80 & .07 & .23 & .00 & -.11 & .68 & .54 & .80 & .20 & -.05 & .79 & .31 & .54 & .07 & -.21 & .78 & .08 & .19 & -.57 & -.61 & .66 & .38 & .86 & .25 & .00 & .86 & .04 & .91 & .02 & -.02 & .99 \\
 & 70 & .08 & .24 & .01 & -.11 & .66 & .64 & .85 & .29 & .00 & .78 & .41 & .65 & .17 & -.11 & .78 & .08 & .22 & -.56 & -.58 & .70 & .34 & .81 & .22 & -.05 & .83 & .05 & .90 & .03 & -.03 & .99 \\
 & 60 & .08 & .23 & .01 & -.12 & .63 & .57 & .82 & .23 & -.03 & .80 & .41 & .62 & .17 & -.14 & .79 & .09 & .22 & -.56 & -.57 & .68 & .29 & .74 & .17 & -.12 & .81 & .05 & .88 & .03 & -.05 & .99 \\
 & 50 & .09 & .23 & .02 & -.11 & .64 & .51 & .76 & .16 & -.08 & .78 & .41 & .62 & .17 & -.14 & .80 & .09 & .23 & -.56 & -.57 & .69 & .27 & .71 & .14 & -.16 & .80 & .06 & .86 & .04 & -.07 & .98 \\
 & 40 & .11 & .24 & .04 & -.10 & .61 & .51 & .72 & .16 & -.12 & .74 & .41 & .59 & .17 & -.17 & .80 & .10 & .27 & -.55 & -.52 & .71 & .21 & .63 & .09 & -.23 & .80 & .07 & .84 & .05 & -.09 & .97 \\
 & 30 & .14 & .25 & .07 & -.10 & .60 & .50 & .70 & .15 & -.15 & .71 & .44 & .61 & .19 & -.15 & .79 & .11 & .31 & -.53 & -.49 & .72 & .16 & .53 & .04 & -.33 & .79 & .07 & .80 & .05 & -.13 & .96 \\
 & 20 & .19 & .29 & .12 & -.06 & .57 & .64 & .76 & .29 & -.08 & .66 & .47 & .62 & .22 & -.13 & .77 & .11 & .29 & -.54 & -.50 & .71 & .11 & .35 & -.01 & -.52 & .71 & .09 & .73 & .08 & -.20 & .93 \\
 & 10 & .45 & .48 & .38 & .14 & .52 & .80 & .84 & .45 & -.01 & .58 & .55 & .63 & .31 & -.13 & .66 & .12 & .32 & -.53 & -.47 & .77 & .00 & .05 & -.12 & -.82 & .71 & .17 & .59 & .15 & -.34 & .84 \\
\midrule
\multirow[t]{11}{*}{WebP} & - & .07 & .34 & .00 & .00 & .75 & .35 & .84 & .00 & .00 & .93 & .25 & .76 & .00 & .00 & .90 & .64 & .80 & .00 & .00 & .62 & .12 & .87 & .00 & .00 & .95 & .02 & .93 & .00 & .00 & 1.0 \\
 & 100 & .05 & .27 & -.02 & -.07 & .73 & .68 & .92 & .34 & .07 & .86 & .31 & .66 & .06 & -.10 & .84 & .09 & .32 & -.56 & -.48 & .78 & .20 & .86 & .08 & -.01 & .91 & .03 & .94 & .01 & .00 & 1.0 \\
 & 90 & .04 & .20 & -.03 & -.14 & .71 & .83 & .96 & .49 & .11 & .83 & .24 & .44 & -.01 & -.32 & .72 & .12 & .41 & -.52 & -.39 & .79 & .19 & .76 & .07 & -.10 & .87 & .03 & .93 & .01 & .00 & 1.0 \\
 & 80 & .03 & .14 & -.04 & -.21 & .66 & .90 & .97 & .56 & .13 & .79 & .17 & .26 & -.07 & -.50 & .64 & .11 & .33 & -.53 & -.47 & .71 & .07 & .44 & -.05 & -.42 & .82 & .05 & .91 & .03 & -.02 & .99 \\
 & 70 & .03 & .13 & -.04 & -.22 & .64 & .93 & .98 & .59 & .14 & .76 & .18 & .25 & -.06 & -.50 & .61 & .11 & .28 & -.53 & -.52 & .67 & .04 & .29 & -.09 & -.58 & .79 & .06 & .89 & .05 & -.04 & .99 \\
 & 60 & .04 & .12 & -.03 & -.22 & .63 & .94 & .98 & .59 & .14 & .76 & .20 & .26 & -.04 & -.50 & .59 & .12 & .28 & -.52 & -.51 & .66 & .03 & .22 & -.10 & -.64 & .78 & .07 & .89 & .05 & -.05 & .99 \\
 & 50 & .04 & .12 & -.03 & -.22 & .62 & .94 & .98 & .59 & .14 & .76 & .22 & .28 & -.02 & -.48 & .59 & .12 & .28 & -.52 & -.51 & .66 & .02 & .16 & -.11 & -.70 & .77 & .08 & .87 & .07 & -.06 & .98 \\
 & 40 & .04 & .13 & -.03 & -.22 & .62 & .94 & .98 & .60 & .14 & .75 & .25 & .31 & .01 & -.45 & .59 & .13 & .28 & -.51 & -.51 & .65 & .01 & .11 & -.11 & -.75 & .75 & .10 & .86 & .08 & -.07 & .97 \\
 & 30 & .05 & .13 & -.02 & -.21 & .60 & .94 & .98 & .59 & .14 & .75 & .29 & .34 & .04 & -.42 & .58 & .14 & .30 & -.50 & -.50 & .65 & .01 & .07 & -.12 & -.79 & .72 & .12 & .84 & .10 & -.10 & .96 \\
 & 20 & .06 & .14 & -.01 & -.20 & .60 & .94 & .98 & .59 & .13 & .74 & .32 & .37 & .08 & -.39 & .58 & .16 & .32 & -.49 & -.48 & .65 & .01 & .05 & -.12 & -.82 & .68 & .14 & .80 & .13 & -.13 & .94 \\
 & 10 & .08 & .16 & .01 & -.18 & .58 & .92 & .97 & .57 & .12 & .73 & .35 & .40 & .11 & -.36 & .58 & .20 & .35 & -.45 & -.45 & .63 & .01 & .03 & -.11 & -.83 & .64 & .18 & .74 & .16 & -.19 & .90 \\
\midrule
\multirow[t]{11}{\linewidth}{Bilinear Upscaling} & 1.00 & .07 & .34 & .00 & .00 & .75 & .35 & .84 & .00 & .00 & .93 & .25 & .76 & .00 & .00 & .90 & .64 & .80 & .00 & .00 & .62 & .12 & .87 & .00 & .00 & .95 & .02 & .93 & .00 & .00 & 1.0 \\
 & 1.10 & .09 & .33 & .02 & -.01 & .72 & .46 & .84 & .11 & -.01 & .88 & .27 & .66 & .03 & -.10 & .88 & .35 & .65 & -.29 & -.14 & .76 & .22 & .93 & .09 & .06 & .95 & .02 & .92 & .00 & -.01 & 1.0 \\
 & 1.20 & .10 & .35 & .03 & .01 & .72 & .51 & .83 & .16 & -.01 & .86 & .26 & .62 & .01 & -.14 & .89 & .40 & .72 & -.24 & -.07 & .78 & .26 & .93 & .14 & .07 & .94 & .02 & .94 & .00 & .01 & 1.0 \\
 & 1.30 & .10 & .38 & .03 & .04 & .74 & .56 & .85 & .21 & .00 & .85 & .27 & .62 & .03 & -.14 & .89 & .22 & .41 & -.42 & -.39 & .65 & .29 & .94 & .17 & .07 & .94 & .02 & .95 & .00 & .02 & 1.0 \\
 & 1.40 & .11 & .40 & .04 & .06 & .74 & .60 & .88 & .25 & .04 & .87 & .30 & .68 & .05 & -.08 & .91 & .35 & .61 & -.30 & -.18 & .74 & .30 & .93 & .18 & .07 & .93 & .03 & .95 & .01 & .02 & 1.0 \\
 & 1.50 & .12 & .41 & .05 & .06 & .74 & .59 & .87 & .25 & .03 & .86 & .30 & .66 & .05 & -.10 & .91 & .50 & .75 & -.14 & -.05 & .74 & .38 & .94 & .25 & .08 & .91 & .02 & .95 & .00 & .02 & 1.0 \\
 & 1.60 & .13 & .42 & .06 & .08 & .74 & .60 & .87 & .25 & .02 & .85 & .30 & .66 & .05 & -.10 & .92 & .27 & .34 & -.38 & -.46 & .60 & .40 & .94 & .28 & .07 & .90 & .03 & .95 & .01 & .02 & 1.0 \\
 & 1.70 & .15 & .44 & .08 & .10 & .74 & .60 & .86 & .25 & .02 & .85 & .30 & .65 & .05 & -.11 & .92 & .36 & .58 & -.29 & -.21 & .70 & .42 & .93 & .30 & .06 & .89 & .03 & .95 & .01 & .02 & 1.0 \\
 & 1.80 & .16 & .46 & .09 & .12 & .74 & .59 & .85 & .24 & .01 & .85 & .30 & .63 & .05 & -.12 & .92 & .36 & .53 & -.29 & -.27 & .65 & .45 & .92 & .33 & .06 & .87 & .03 & .94 & .01 & .01 & 1.0 \\
 & 1.90 & .17 & .48 & .10 & .13 & .75 & .59 & .85 & .24 & .00 & .84 & .30 & .64 & .06 & -.12 & .93 & .45 & .59 & -.20 & -.21 & .63 & .50 & .92 & .37 & .06 & .86 & .04 & .94 & .02 & .01 & 1.0 \\
 & 2.00 & .17 & .47 & .10 & .12 & .73 & .58 & .84 & .23 & -.01 & .84 & .30 & .63 & .05 & -.13 & .93 & .31 & .29 & -.33 & -.50 & .54 & .65 & .96 & .53 & .10 & .81 & .04 & .93 & .02 & .00 & 1.0 \\
\midrule
\multirow[t]{11}{\linewidth}{Bilinear Downscaling} & 1.00 & .07 & .34 & .00 & .00 & .75 & .35 & .84 & .00 & .00 & .93 & .25 & .76 & .00 & .00 & .90 & .64 & .80 & .00 & .00 & .62 & .12 & .87 & .00 & .00 & .95 & .02 & .93 & .00 & .00 & 1.0 \\
 & 0.95 & .07 & .28 & .01 & -.06 & .70 & .40 & .82 & .06 & -.02 & .90 & .26 & .68 & .02 & -.08 & .88 & .55 & .81 & -.10 & .01 & .77 & .20 & .91 & .07 & .04 & .95 & .02 & .91 & .00 & -.02 & 1.0 \\
 & 0.90 & .07 & .24 & .00 & -.10 & .68 & .41 & .83 & .07 & -.01 & .90 & .25 & .65 & .00 & -.11 & .87 & .58 & .83 & -.07 & .03 & .76 & .20 & .90 & .07 & .04 & .94 & .02 & .90 & .00 & -.04 & .99 \\
 & 0.85 & .06 & .25 & -.01 & -.10 & .68 & .42 & .83 & .08 & -.01 & .89 & .25 & .64 & .01 & -.12 & .87 & .60 & .83 & -.04 & .03 & .74 & .19 & .88 & .07 & .02 & .93 & .02 & .88 & .00 & -.05 & .99 \\
 & 0.80 & .05 & .22 & -.02 & -.12 & .67 & .44 & .84 & .09 & .00 & .90 & .26 & .64 & .02 & -.12 & .86 & .62 & .83 & -.03 & .04 & .74 & .21 & .87 & .09 & .00 & .92 & .02 & .86 & .00 & -.07 & .99 \\
 & 0.75 & .05 & .20 & -.02 & -.14 & .66 & .44 & .85 & .10 & .01 & .90 & .27 & .63 & .03 & -.13 & .86 & .62 & .82 & -.02 & .02 & .72 & .19 & .84 & .06 & -.02 & .91 & .02 & .84 & .00 & -.09 & .98 \\
 & 0.70 & .05 & .19 & -.02 & -.15 & .64 & .44 & .85 & .10 & .01 & .90 & .29 & .63 & .05 & -.12 & .84 & .65 & .83 & .01 & .04 & .71 & .17 & .82 & .05 & -.05 & .90 & .02 & .81 & .00 & -.12 & .97 \\
 & 0.65 & .04 & .17 & -.03 & -.17 & .63 & .46 & .86 & .11 & .01 & .90 & .32 & .63 & .07 & -.12 & .82 & .65 & .83 & .01 & .03 & .71 & .16 & .77 & .03 & -.09 & .89 & .02 & .78 & .01 & -.15 & .96 \\
 & 0.60 & .04 & .16 & -.03 & -.18 & .63 & .47 & .86 & .12 & .01 & .90 & .35 & .64 & .11 & -.11 & .80 & .67 & .83 & .02 & .03 & .69 & .14 & .73 & .01 & -.14 & .88 & .04 & .74 & .02 & -.19 & .93 \\
 & 0.55 & .03 & .15 & -.03 & -.20 & .61 & .46 & .85 & .11 & .00 & .90 & .36 & .63 & .12 & -.13 & .78 & .65 & .81 & .01 & .02 & .68 & .14 & .69 & .01 & -.18 & .86 & .05 & .70 & .03 & -.23 & .91 \\
 & 0.50 & .03 & .13 & -.04 & -.21 & .59 & .45 & .82 & .10 & -.02 & .88 & .36 & .59 & .11 & -.17 & .75 & .61 & .74 & -.03 & -.06 & .65 & .14 & .63 & .01 & -.23 & .83 & .05 & .65 & .04 & -.28 & .88 \\
\midrule
\multirow[t]{8}{*}{Rotation} & 0 & .07 & .34 & .00 & .00 & .75 & .35 & .84 & .00 & .00 & .93 & .25 & .76 & .00 & .00 & .90 & .64 & .80 & .00 & .00 & .62 & .12 & .87 & .00 & .00 & .95 & .02 & .93 & .00 & .00 & 1.0 \\
 & 45 & .31 & .62 & .24 & .28 & .73 & .45 & .74 & .10 & -.10 & .80 & .49 & .78 & .24 & .02 & .83 & .32 & .57 & -.32 & -.22 & .76 & .04 & .55 & -.09 & -.31 & .92 & .12 & .82 & .10 & -.11 & .94 \\
 & 90 & .46 & .50 & .39 & .16 & .53 & .73 & .85 & .38 & .00 & .71 & .55 & .74 & .30 & -.01 & .74 & .24 & .71 & -.40 & -.08 & .88 & .03 & .83 & -.09 & -.04 & .97 & .34 & .93 & .32 & .00 & .95 \\
 & 135 & .54 & .79 & .47 & .45 & .71 & .53 & .80 & .18 & -.05 & .80 & .58 & .81 & .34 & .05 & .79 & .39 & .58 & -.25 & -.22 & .71 & .04 & .55 & -.09 & -.32 & .92 & .19 & .82 & .17 & -.11 & .92 \\
 & 180 & .26 & .59 & .19 & .25 & .73 & .39 & .85 & .04 & .01 & .93 & .34 & .78 & .10 & .02 & .88 & .68 & .79 & .04 & .00 & .55 & .13 & .87 & .00 & .00 & .95 & .05 & .94 & .03 & .01 & .99 \\
 & 225 & .55 & .79 & .48 & .45 & .71 & .53 & .79 & .18 & -.06 & .79 & .58 & .81 & .34 & .05 & .79 & .40 & .60 & -.24 & -.20 & .72 & .03 & .54 & -.09 & -.33 & .92 & .20 & .82 & .18 & -.11 & .92 \\
 & 270 & .48 & .50 & .41 & .16 & .51 & .72 & .85 & .37 & .00 & .72 & .55 & .75 & .31 & -.01 & .73 & .54 & .72 & -.11 & -.08 & .70 & .04 & .82 & -.09 & -.04 & .97 & .36 & .93 & .34 & .00 & .95 \\
 & 315 & .31 & .61 & .24 & .27 & .73 & .45 & .75 & .10 & -.10 & .81 & .48 & .78 & .24 & .02 & .84 & .33 & .58 & -.32 & -.22 & .76 & .03 & .54 & -.09 & -.33 & .92 & .11 & .81 & .09 & -.12 & .94 \\
\midrule
\multirow[t]{2}{\linewidth}{Grayscale Conversion} & RGB & .07 & .34 & .00 & .00 & .75 & .35 & .84 & .00 & .00 & .93 & .25 & .76 & .00 & .00 & .90 & .64 & .80 & .00 & .00 & .62 & .12 & .87 & .00 & .00 & .95 & .02 & .93 & .00 & .00 & 1.0 \\
 & Gray & .06 & .23 & -.01 & -.11 & .69 & .51 & .79 & .16 & -.05 & .81 & .16 & .42 & -.09 & -.34 & .84 & .68 & .81 & .03 & .01 & .58 & .00 & .00 & -.12 & -.86 & .68 & .04 & .92 & .02 & -.01 & 1.0 \\
\midrule
\multirow[t]{11}{\linewidth}{Bicubic Upscaling} & 1.00 & .07 & .34 & .00 & .00 & .75 & .35 & .84 & .00 & .00 & .93 & .25 & .76 & .00 & .00 & .90 & .64 & .80 & .00 & .00 & .62 & .12 & .87 & .00 & .00 & .95 & .02 & .93 & .00 & .00 & 1.0 \\
 & 1.10 & .09 & .33 & .02 & -.01 & .72 & .31 & .78 & -.03 & -.07 & .91 & .22 & .62 & -.02 & -.13 & .87 & .37 & .64 & -.28 & -.15 & .74 & .17 & .92 & .05 & .06 & .96 & .01 & .91 & .00 & -.02 & 1.0 \\
 & 1.20 & .09 & .35 & .02 & .01 & .73 & .35 & .74 & .00 & -.10 & .87 & .22 & .59 & -.02 & -.17 & .87 & .34 & .66 & -.30 & -.14 & .77 & .20 & .93 & .08 & .07 & .96 & .02 & .93 & .00 & .00 & 1.0 \\
 & 1.30 & .10 & .38 & .03 & .04 & .74 & .39 & .75 & .04 & -.09 & .86 & .24 & .60 & -.01 & -.16 & .88 & .17 & .34 & -.48 & -.46 & .65 & .24 & .94 & .12 & .07 & .95 & .02 & .94 & .00 & .01 & 1.0 \\
 & 1.40 & .11 & .40 & .04 & .06 & .75 & .41 & .78 & .06 & -.06 & .87 & .26 & .64 & .01 & -.12 & .90 & .20 & .44 & -.44 & -.35 & .73 & .28 & .94 & .16 & .07 & .94 & .02 & .94 & .00 & .00 & 1.0 \\
 & 1.50 & .12 & .43 & .05 & .08 & .75 & .41 & .77 & .07 & -.07 & .86 & .26 & .64 & .02 & -.12 & .90 & .36 & .63 & -.29 & -.17 & .73 & .31 & .94 & .18 & .07 & .93 & .02 & .93 & .00 & .00 & 1.0 \\
 & 1.60 & .13 & .44 & .06 & .10 & .75 & .41 & .75 & .07 & -.09 & .85 & .27 & .64 & .02 & -.12 & .91 & .18 & .29 & -.46 & -.50 & .62 & .34 & .94 & .22 & .08 & .92 & .02 & .93 & .00 & .00 & 1.0 \\
 & 1.70 & .14 & .46 & .07 & .11 & .75 & .42 & .75 & .07 & -.10 & .84 & .27 & .64 & .03 & -.12 & .91 & .26 & .51 & -.39 & -.29 & .73 & .40 & .94 & .28 & .07 & .90 & .02 & .93 & .00 & .00 & 1.0 \\
 & 1.80 & .15 & .47 & .08 & .13 & .75 & .42 & .74 & .07 & -.10 & .83 & .28 & .64 & .03 & -.12 & .91 & .27 & .47 & -.38 & -.33 & .69 & .46 & .94 & .34 & .08 & .89 & .02 & .92 & .00 & -.01 & 1.0 \\
 & 1.90 & .16 & .49 & .09 & .15 & .76 & .43 & .73 & .09 & -.11 & .82 & .28 & .64 & .03 & -.12 & .93 & .33 & .48 & -.32 & -.31 & .64 & .53 & .94 & .41 & .08 & .86 & .02 & .92 & .00 & -.01 & 1.0 \\
 & 2.00 & .16 & .50 & .09 & .15 & .76 & .45 & .73 & .10 & -.11 & .81 & .29 & .64 & .04 & -.12 & .93 & .25 & .21 & -.40 & -.59 & .51 & .58 & .96 & .46 & .10 & .85 & .03 & .92 & .01 & -.01 & 1.0 \\
\midrule
\multirow[t]{11}{\linewidth}{Bicubic Downscaling} & 1.00 & .07 & .34 & .00 & .00 & .75 & .35 & .84 & .00 & .00 & .93 & .25 & .76 & .00 & .00 & .90 & .64 & .80 & .00 & .00 & .62 & .12 & .87 & .00 & .00 & .95 & .02 & .93 & .00 & .00 & 1.0 \\
 & 0.95 & .07 & .27 & .00 & -.07 & .70 & .30 & .77 & -.05 & -.07 & .91 & .21 & .63 & -.03 & -.12 & .87 & .59 & .78 & -.05 & -.02 & .68 & .15 & .89 & .03 & .02 & .95 & .02 & .90 & .00 & -.03 & 1.0 \\
 & 0.90 & .07 & .24 & .00 & -.10 & .68 & .30 & .77 & -.05 & -.07 & .91 & .21 & .61 & -.04 & -.15 & .85 & .61 & .79 & -.03 & -.01 & .67 & .15 & .87 & .02 & .01 & .95 & .02 & .89 & .00 & -.04 & .99 \\
 & 0.85 & .06 & .25 & -.01 & -.10 & .69 & .30 & .76 & -.05 & -.09 & .90 & .21 & .60 & -.03 & -.16 & .84 & .62 & .79 & -.02 & -.01 & .66 & .13 & .84 & .01 & -.02 & .94 & .02 & .87 & .00 & -.06 & .99 \\
 & 0.80 & .05 & .23 & -.02 & -.12 & .68 & .30 & .75 & -.05 & -.10 & .90 & .23 & .59 & -.02 & -.17 & .84 & .61 & .78 & -.03 & -.01 & .66 & .18 & .82 & .05 & -.04 & .91 & .02 & .84 & .00 & -.09 & .99 \\
 & 0.75 & .05 & .20 & -.02 & -.14 & .66 & .30 & .75 & -.05 & -.10 & .90 & .24 & .58 & .00 & -.18 & .83 & .63 & .77 & -.01 & -.03 & .63 & .13 & .78 & .01 & -.08 & .91 & .02 & .82 & .00 & -.11 & .98 \\
 & 0.70 & .04 & .19 & -.02 & -.15 & .65 & .29 & .74 & -.06 & -.11 & .90 & .26 & .57 & .01 & -.18 & .81 & .63 & .79 & -.01 & -.01 & .64 & .12 & .75 & -.01 & -.12 & .90 & .02 & .79 & .00 & -.14 & .97 \\
 & 0.65 & .04 & .17 & -.03 & -.17 & .64 & .29 & .73 & -.06 & -.12 & .89 & .28 & .56 & .03 & -.19 & .79 & .62 & .78 & -.02 & -.02 & .63 & .10 & .70 & -.02 & -.17 & .88 & .02 & .76 & .00 & -.18 & .96 \\
 & 0.60 & .04 & .16 & -.03 & -.19 & .63 & .29 & .71 & -.06 & -.13 & .88 & .30 & .56 & .06 & -.20 & .77 & .64 & .78 & -.01 & -.02 & .62 & .10 & .66 & -.03 & -.21 & .87 & .04 & .72 & .02 & -.21 & .93 \\
 & 0.55 & .03 & .14 & -.04 & -.20 & .62 & .28 & .70 & -.07 & -.14 & .88 & .32 & .55 & .08 & -.20 & .74 & .63 & .77 & -.01 & -.03 & .62 & .08 & .60 & -.05 & -.27 & .86 & .05 & .67 & .03 & -.26 & .90 \\
 & 0.50 & .03 & .13 & -.04 & -.21 & .61 & .27 & .68 & -.07 & -.16 & .87 & .34 & .55 & .10 & -.20 & .72 & .63 & .70 & -.02 & -.10 & .57 & .08 & .53 & -.05 & -.33 & .83 & .05 & .62 & .03 & -.31 & .88 \\
\bottomrule
\end{tabular}

\end{table}

\begin{table}[htbp]
    \centering
    \fontsize{5}{5}\selectfont
    \setlength{\tabcolsep}{1.05pt}
    \caption{\textbf{Numerical values of the plots for CLIC2020\,/\,SynthCLIC.}}
    \label{tab:clic2020_all_vs_synthclic_all_plot_values}
    \begin{tabular}{@{\ }p{30pt}rrrrrrrrrrrrrrrrrrrrrrrrrrrrrrr@{\ }}
\toprule
 & & \multicolumn{5}{c}{UnivFD} & \multicolumn{5}{c}{DRCT} & \multicolumn{5}{c}{RINE} & \multicolumn{5}{c}{AIDE} & \multicolumn{5}{c}{SPAI} & \multicolumn{5}{c}{B-Free} \\
 \cmidrule(l{1pt}r{1pt}){3-7} \cmidrule(l{1pt}r{1pt}){8-12} \cmidrule(l{1pt}r{1pt}){13-17} \cmidrule(l{1pt}r{1pt}){18-22} \cmidrule(l{1pt}r{1pt}){23-27} \cmidrule(l{1pt}r{1pt}){28-32}
 & & \multicolumn{2}{c}{Score} & \multicolumn{2}{c}{MSS} & AUC & \multicolumn{2}{c}{Score} & \multicolumn{2}{c}{MSS} & AUC & \multicolumn{2}{c}{Score} & \multicolumn{2}{c}{MSS} & AUC & \multicolumn{2}{c}{Score} & \multicolumn{2}{c}{MSS} & AUC & \multicolumn{2}{c}{Score} & \multicolumn{2}{c}{MSS} & AUC & \multicolumn{2}{c}{Score} & \multicolumn{2}{c}{MSS} & AUC \\
\cmidrule(r{1pt}){3-4} \cmidrule(l{1pt}){5-6}
\cmidrule(r{1pt}){8-9} \cmidrule(l{1pt}){10-11}
\cmidrule(r{1pt}){13-14} \cmidrule(l{1pt}){15-16}
\cmidrule(r{1pt}){18-19} \cmidrule(l{1pt}){20-21}
\cmidrule(r{1pt}){23-24} \cmidrule(l{1pt}){25-26}
\cmidrule(r{1pt}){28-29} \cmidrule(l{1pt}){30-31}
 Transform & Level & Real & Fake & Real & Fake &  & Real & Fake & Real & Fake &  & Real & Fake & Real & Fake &  & Real & Fake & Real & Fake &  & Real & Fake & Real & Fake &  & Real & Fake & Real & Fake &  \\
\midrule
\multirow[t]{11}{*}{JPEG} & - & .06 & .06 & .00 & .00 & .54 & .79 & .80 & .00 & .00 & .51 & .25 & .64 & .00 & .00 & .88 & .56 & .86 & .00 & .00 & .81 & .29 & .80 & .00 & .00 & .84 & .04 & .50 & .00 & .00 & .83 \\
 & 100 & .06 & .06 & .00 & .00 & .53 & .86 & .83 & .07 & .04 & .50 & .30 & .64 & .05 & .01 & .85 & .25 & .27 & -.30 & -.59 & .51 & .31 & .80 & .02 & .00 & .84 & .05 & .51 & .00 & .01 & .83 \\
 & 90 & .04 & .04 & -.01 & -.02 & .50 & .69 & .65 & -.10 & -.14 & .48 & .28 & .39 & .03 & -.25 & .67 & .09 & .12 & -.46 & -.74 & .62 & .42 & .81 & .13 & .01 & .79 & .05 & .46 & .01 & -.04 & .79 \\
 & 80 & .06 & .05 & .00 & -.01 & .49 & .77 & .78 & -.02 & -.02 & .52 & .32 & .42 & .08 & -.22 & .64 & .13 & .22 & -.43 & -.64 & .66 & .46 & .77 & .17 & -.03 & .73 & .06 & .43 & .02 & -.07 & .77 \\
 & 70 & .08 & .05 & .02 & -.01 & .46 & .83 & .84 & .04 & .04 & .54 & .43 & .50 & .18 & -.14 & .62 & .16 & .28 & -.40 & -.58 & .67 & .43 & .69 & .15 & -.10 & .69 & .07 & .40 & .03 & -.10 & .74 \\
 & 60 & .08 & .05 & .02 & -.01 & .44 & .77 & .80 & -.02 & .01 & .56 & .42 & .49 & .17 & -.15 & .63 & .17 & .29 & -.39 & -.57 & .66 & .37 & .59 & .08 & -.21 & .66 & .08 & .37 & .04 & -.13 & .72 \\
 & 50 & .09 & .05 & .04 & -.01 & .43 & .71 & .74 & -.08 & -.05 & .55 & .43 & .52 & .18 & -.12 & .65 & .18 & .30 & -.38 & -.56 & .66 & .36 & .56 & .07 & -.24 & .65 & .08 & .34 & .04 & -.16 & .70 \\
 & 40 & .12 & .06 & .06 & .00 & .42 & .68 & .69 & -.11 & -.11 & .51 & .44 & .53 & .19 & -.11 & .67 & .22 & .35 & -.34 & -.51 & .66 & .30 & .47 & .01 & -.33 & .63 & .09 & .31 & .05 & -.19 & .67 \\
 & 30 & .14 & .07 & .08 & .01 & .40 & .66 & .66 & -.13 & -.14 & .49 & .46 & .55 & .21 & -.08 & .67 & .24 & .39 & -.31 & -.47 & .66 & .25 & .37 & -.04 & -.42 & .61 & .09 & .27 & .05 & -.24 & .64 \\
 & 20 & .22 & .11 & .16 & .05 & .38 & .76 & .73 & -.03 & -.07 & .46 & .49 & .57 & .25 & -.07 & .64 & .24 & .38 & -.32 & -.48 & .66 & .17 & .24 & -.12 & -.56 & .57 & .10 & .21 & .06 & -.29 & .58 \\
 & 10 & .52 & .30 & .46 & .23 & .33 & .86 & .83 & .07 & .03 & .45 & .53 & .58 & .28 & -.06 & .61 & .25 & .37 & -.31 & -.49 & .65 & .01 & .02 & -.28 & -.78 & .57 & .16 & .18 & .12 & -.33 & .47 \\
\midrule
\multirow[t]{11}{*}{WebP} & - & .06 & .06 & .00 & .00 & .54 & .79 & .80 & .00 & .00 & .51 & .25 & .64 & .00 & .00 & .88 & .56 & .86 & .00 & .00 & .81 & .29 & .80 & .00 & .00 & .84 & .04 & .50 & .00 & .00 & .83 \\
 & 100 & .05 & .04 & -.01 & -.02 & .51 & .90 & .91 & .11 & .11 & .56 & .30 & .51 & .06 & -.13 & .76 & .14 & .18 & -.42 & -.68 & .62 & .32 & .77 & .03 & -.03 & .81 & .05 & .50 & .01 & .00 & .83 \\
 & 90 & .04 & .03 & -.02 & -.04 & .47 & .94 & .95 & .15 & .16 & .58 & .22 & .31 & -.03 & -.33 & .64 & .24 & .40 & -.32 & -.46 & .68 & .23 & .60 & -.06 & -.19 & .77 & .06 & .47 & .01 & -.03 & .80 \\
 & 80 & .03 & .02 & -.02 & -.04 & .44 & .97 & .97 & .18 & .18 & .55 & .17 & .18 & -.08 & -.46 & .51 & .21 & .34 & -.35 & -.52 & .66 & .08 & .24 & -.21 & -.56 & .71 & .08 & .44 & .04 & -.06 & .76 \\
 & 70 & .04 & .01 & -.02 & -.05 & .42 & .97 & .98 & .18 & .18 & .56 & .18 & .17 & -.07 & -.46 & .48 & .21 & .34 & -.34 & -.52 & .65 & .04 & .13 & -.24 & -.66 & .68 & .10 & .41 & .06 & -.09 & .73 \\
 & 60 & .04 & .01 & -.02 & -.05 & .40 & .97 & .98 & .18 & .18 & .55 & .19 & .18 & -.05 & -.46 & .46 & .22 & .35 & -.34 & -.51 & .64 & .03 & .09 & -.26 & -.70 & .66 & .11 & .40 & .07 & -.11 & .72 \\
 & 50 & .04 & .02 & -.02 & -.05 & .40 & .97 & .98 & .18 & .18 & .55 & .21 & .20 & -.04 & -.44 & .46 & .23 & .35 & -.33 & -.51 & .64 & .02 & .06 & -.26 & -.73 & .64 & .12 & .38 & .08 & -.12 & .70 \\
 & 40 & .05 & .02 & -.01 & -.04 & .38 & .97 & .98 & .18 & .18 & .56 & .23 & .21 & -.02 & -.42 & .46 & .24 & .36 & -.32 & -.50 & .63 & .02 & .04 & -.27 & -.75 & .61 & .13 & .36 & .09 & -.14 & .67 \\
 & 30 & .06 & .02 & .00 & -.04 & .37 & .97 & .97 & .18 & .18 & .56 & .24 & .24 & -.01 & -.40 & .48 & .26 & .38 & -.30 & -.48 & .63 & .02 & .03 & -.27 & -.77 & .58 & .15 & .33 & .11 & -.17 & .64 \\
 & 20 & .08 & .02 & .02 & -.04 & .36 & .96 & .97 & .17 & .17 & .55 & .26 & .26 & .01 & -.38 & .50 & .29 & .40 & -.27 & -.46 & .61 & .01 & .02 & -.28 & -.78 & .55 & .17 & .30 & .12 & -.20 & .60 \\
 & 10 & .11 & .04 & .05 & -.02 & .34 & .95 & .96 & .16 & .16 & .55 & .28 & .29 & .03 & -.35 & .52 & .34 & .43 & -.22 & -.43 & .59 & .02 & .02 & -.27 & -.78 & .51 & .19 & .27 & .15 & -.24 & .55 \\
\midrule
\multirow[t]{11}{\linewidth}{Bilinear Upscaling} & 1.00 & .06 & .06 & .00 & .00 & .54 & .79 & .80 & .00 & .00 & .51 & .25 & .64 & .00 & .00 & .88 & .56 & .86 & .00 & .00 & .81 & .29 & .80 & .00 & .00 & .84 & .04 & .50 & .00 & .00 & .83 \\
 & 1.10 & .08 & .07 & .02 & .01 & .49 & .77 & .79 & -.02 & .00 & .52 & .25 & .55 & .00 & -.08 & .86 & .41 & .65 & -.15 & -.21 & .72 & .35 & .86 & .06 & .06 & .87 & .04 & .51 & .00 & .01 & .83 \\
 & 1.20 & .10 & .08 & .04 & .02 & .49 & .79 & .81 & .00 & .01 & .52 & .25 & .55 & .00 & -.08 & .86 & .50 & .72 & -.06 & -.14 & .71 & .37 & .89 & .08 & .09 & .88 & .05 & .57 & .01 & .07 & .88 \\
 & 1.30 & .11 & .10 & .05 & .03 & .49 & .81 & .84 & .02 & .04 & .54 & .26 & .59 & .02 & -.05 & .88 & .30 & .44 & -.26 & -.42 & .61 & .39 & .92 & .11 & .12 & .89 & .06 & .63 & .02 & .13 & .91 \\
 & 1.40 & .13 & .11 & .07 & .05 & .50 & .82 & .86 & .03 & .06 & .55 & .29 & .62 & .04 & -.02 & .88 & .45 & .68 & -.11 & -.18 & .71 & .40 & .93 & .12 & .13 & .90 & .06 & .67 & .02 & .17 & .92 \\
 & 1.50 & .14 & .12 & .08 & .06 & .50 & .81 & .85 & .02 & .05 & .55 & .29 & .60 & .04 & -.03 & .88 & .62 & .85 & .06 & -.01 & .76 & .47 & .95 & .18 & .16 & .90 & .06 & .70 & .02 & .20 & .94 \\
 & 1.60 & .16 & .14 & .10 & .08 & .49 & .81 & .84 & .02 & .05 & .55 & .30 & .61 & .05 & -.03 & .88 & .27 & .49 & -.29 & -.37 & .71 & .47 & .96 & .19 & .16 & .90 & .08 & .73 & .04 & .23 & .95 \\
 & 1.70 & .18 & .15 & .12 & .09 & .49 & .81 & .84 & .02 & .04 & .55 & .30 & .60 & .05 & -.03 & .88 & .41 & .70 & -.15 & -.16 & .75 & .48 & .96 & .20 & .16 & .89 & .09 & .75 & .05 & .25 & .95 \\
 & 1.80 & .19 & .17 & .13 & .11 & .49 & .80 & .83 & .01 & .04 & .55 & .30 & .59 & .05 & -.04 & .88 & .38 & .69 & -.18 & -.17 & .76 & .51 & .96 & .22 & .17 & .89 & .10 & .77 & .06 & .27 & .95 \\
 & 1.90 & .20 & .18 & .14 & .12 & .49 & .79 & .84 & .00 & .04 & .56 & .31 & .59 & .06 & -.04 & .87 & .46 & .76 & -.10 & -.10 & .76 & .53 & .96 & .24 & .17 & .89 & .11 & .79 & .07 & .29 & .96 \\
 & 2.00 & .21 & .19 & .15 & .12 & .49 & .78 & .81 & -.01 & .02 & .54 & .30 & .58 & .05 & -.06 & .87 & .21 & .45 & -.34 & -.41 & .74 & .64 & .98 & .35 & .18 & .86 & .11 & .81 & .07 & .31 & .97 \\
\midrule
\multirow[t]{11}{\linewidth}{Bilinear Downscaling} & 1.00 & .06 & .06 & .00 & .00 & .54 & .79 & .80 & .00 & .00 & .51 & .25 & .64 & .00 & .00 & .88 & .56 & .86 & .00 & .00 & .81 & .29 & .80 & .00 & .00 & .84 & .04 & .50 & .00 & .00 & .83 \\
 & 0.95 & .06 & .05 & .01 & -.01 & .49 & .76 & .78 & -.03 & -.02 & .52 & .25 & .55 & .00 & -.09 & .84 & .59 & .78 & .03 & -.08 & .70 & .34 & .82 & .05 & .02 & .84 & .04 & .44 & .00 & -.06 & .78 \\
 & 0.90 & .05 & .04 & .00 & -.02 & .48 & .76 & .78 & -.03 & -.02 & .53 & .23 & .51 & -.02 & -.13 & .83 & .62 & .77 & .06 & -.09 & .67 & .34 & .80 & .05 & .00 & .83 & .03 & .40 & -.01 & -.10 & .77 \\
 & 0.85 & .05 & .03 & -.01 & -.03 & .48 & .76 & .79 & -.03 & -.01 & .53 & .23 & .50 & -.01 & -.14 & .82 & .65 & .78 & .09 & -.08 & .65 & .32 & .77 & .04 & -.03 & .81 & .03 & .36 & -.01 & -.14 & .75 \\
 & 0.80 & .05 & .03 & -.01 & -.03 & .47 & .76 & .80 & -.03 & .00 & .54 & .24 & .49 & -.01 & -.15 & .80 & .64 & .79 & .08 & -.07 & .67 & .31 & .74 & .03 & -.05 & .79 & .03 & .32 & -.01 & -.18 & .73 \\
 & 0.75 & .04 & .03 & -.02 & -.03 & .47 & .76 & .80 & -.03 & .01 & .55 & .24 & .48 & .00 & -.15 & .79 & .67 & .76 & .11 & -.11 & .60 & .31 & .72 & .02 & -.08 & .78 & .03 & .29 & -.01 & -.21 & .71 \\
 & 0.70 & .04 & .02 & -.02 & -.04 & .47 & .77 & .80 & -.02 & .01 & .54 & .25 & .48 & .01 & -.16 & .77 & .68 & .79 & .13 & -.07 & .63 & .30 & .67 & .01 & -.12 & .75 & .03 & .24 & -.01 & -.26 & .68 \\
 & 0.65 & .03 & .02 & -.02 & -.04 & .47 & .77 & .81 & -.02 & .01 & .54 & .27 & .48 & .02 & -.15 & .76 & .71 & .79 & .15 & -.07 & .60 & .30 & .63 & .01 & -.17 & .72 & .05 & .24 & .01 & -.26 & .65 \\
 & 0.60 & .03 & .02 & -.03 & -.04 & .47 & .78 & .81 & -.01 & .01 & .54 & .29 & .49 & .04 & -.15 & .74 & .73 & .81 & .17 & -.05 & .60 & .29 & .57 & .00 & -.23 & .69 & .06 & .22 & .02 & -.28 & .62 \\
 & 0.55 & .03 & .02 & -.03 & -.05 & .47 & .77 & .80 & -.02 & .00 & .54 & .30 & .49 & .05 & -.15 & .72 & .72 & .81 & .16 & -.05 & .61 & .28 & .52 & .00 & -.28 & .66 & .07 & .20 & .03 & -.30 & .59 \\
 & 0.50 & .02 & .01 & -.03 & -.05 & .47 & .75 & .78 & -.04 & -.02 & .54 & .30 & .47 & .05 & -.17 & .70 & .68 & .78 & .13 & -.08 & .64 & .28 & .46 & -.01 & -.34 & .63 & .07 & .17 & .03 & -.33 & .56 \\
\midrule
\multirow[t]{8}{*}{Rotation} & 0 & .06 & .06 & .00 & .00 & .54 & .79 & .80 & .00 & .00 & .51 & .25 & .64 & .00 & .00 & .88 & .56 & .86 & .00 & .00 & .81 & .29 & .80 & .00 & .00 & .84 & .04 & .50 & .00 & .00 & .83 \\
 & 45 & .24 & .27 & .19 & .21 & .54 & .68 & .65 & -.11 & -.15 & .46 & .42 & .68 & .17 & .04 & .80 & .44 & .57 & -.12 & -.29 & .64 & .10 & .50 & -.19 & -.29 & .79 & .10 & .41 & .06 & -.10 & .73 \\
 & 90 & .19 & .23 & .13 & .17 & .57 & .80 & .83 & .01 & .03 & .55 & .31 & .64 & .06 & .00 & .87 & .49 & .64 & -.07 & -.22 & .65 & .23 & .68 & -.05 & -.11 & .79 & .11 & .59 & .07 & .08 & .85 \\
 & 135 & .42 & .48 & .36 & .42 & .55 & .72 & .70 & -.07 & -.10 & .48 & .46 & .73 & .21 & .09 & .81 & .47 & .57 & -.09 & -.29 & .61 & .10 & .50 & -.19 & -.30 & .79 & .12 & .44 & .08 & -.06 & .73 \\
 & 180 & .19 & .24 & .13 & .18 & .55 & .80 & .82 & .00 & .02 & .53 & .30 & .69 & .05 & .05 & .88 & .60 & .86 & .04 & .00 & .79 & .30 & .80 & .01 & .01 & .85 & .07 & .62 & .03 & .12 & .87 \\
 & 225 & .43 & .49 & .37 & .43 & .55 & .72 & .70 & -.07 & -.10 & .47 & .46 & .73 & .21 & .09 & .81 & .48 & .59 & -.08 & -.27 & .62 & .10 & .50 & -.19 & -.29 & .80 & .13 & .45 & .09 & -.05 & .73 \\
 & 270 & .19 & .24 & .13 & .18 & .57 & .79 & .83 & .00 & .03 & .55 & .31 & .64 & .06 & .00 & .87 & .49 & .62 & -.06 & -.25 & .62 & .22 & .67 & -.06 & -.12 & .79 & .11 & .58 & .07 & .08 & .85 \\
 & 315 & .25 & .26 & .19 & .20 & .53 & .69 & .65 & -.10 & -.14 & .46 & .41 & .68 & .17 & .04 & .80 & .47 & .58 & -.09 & -.29 & .62 & .09 & .48 & -.20 & -.32 & .79 & .08 & .38 & .04 & -.12 & .73 \\
\midrule
\multirow[t]{2}{\linewidth}{Grayscale Conversion} & RGB & .06 & .06 & .00 & .00 & .54 & .79 & .80 & .00 & .00 & .51 & .25 & .64 & .00 & .00 & .88 & .56 & .86 & .00 & .00 & .81 & .29 & .80 & .00 & .00 & .84 & .04 & .50 & .00 & .00 & .83 \\
 & Gray & .07 & .07 & .01 & .00 & .50 & .75 & .73 & -.04 & -.07 & .49 & .16 & .26 & -.09 & -.38 & .69 & .62 & .86 & .06 & .00 & .78 & .00 & .00 & -.29 & -.80 & .68 & .05 & .52 & .01 & .02 & .84 \\
\midrule
\multirow[t]{11}{\linewidth}{Bicubic Upscaling} & 1.00 & .06 & .06 & .00 & .00 & .54 & .79 & .80 & .00 & .00 & .51 & .25 & .64 & .00 & .00 & .88 & .56 & .86 & .00 & .00 & .81 & .29 & .80 & .00 & .00 & .84 & .04 & .50 & .00 & .00 & .83 \\
 & 1.10 & .07 & .07 & .02 & .01 & .52 & .73 & .72 & -.06 & -.07 & .49 & .21 & .54 & -.04 & -.09 & .88 & .38 & .65 & -.18 & -.21 & .74 & .31 & .85 & .02 & .05 & .88 & .03 & .50 & -.01 & .00 & .85 \\
 & 1.20 & .09 & .09 & .03 & .03 & .52 & .74 & .72 & -.05 & -.07 & .47 & .21 & .55 & -.04 & -.08 & .89 & .38 & .65 & -.17 & -.21 & .74 & .33 & .88 & .04 & .09 & .89 & .03 & .56 & -.01 & .06 & .89 \\
 & 1.30 & .10 & .10 & .04 & .04 & .52 & .75 & .75 & -.04 & -.05 & .49 & .22 & .59 & -.02 & -.05 & .90 & .22 & .36 & -.33 & -.50 & .62 & .34 & .91 & .06 & .11 & .90 & .04 & .61 & .00 & .11 & .92 \\
 & 1.40 & .11 & .12 & .05 & .06 & .53 & .75 & .76 & -.04 & -.04 & .50 & .24 & .61 & -.01 & -.02 & .91 & .29 & .53 & -.27 & -.33 & .73 & .37 & .93 & .08 & .13 & .91 & .04 & .65 & .00 & .15 & .93 \\
 & 1.50 & .13 & .13 & .07 & .07 & .53 & .75 & .75 & -.04 & -.05 & .48 & .25 & .61 & .00 & -.03 & .91 & .44 & .76 & -.11 & -.10 & .79 & .40 & .95 & .11 & .15 & .92 & .05 & .68 & .01 & .18 & .95 \\
 & 1.60 & .14 & .14 & .08 & .08 & .52 & .75 & .73 & -.04 & -.06 & .47 & .26 & .61 & .01 & -.02 & .90 & .20 & .44 & -.35 & -.42 & .75 & .43 & .96 & .14 & .17 & .92 & .05 & .72 & .01 & .21 & .96 \\
 & 1.70 & .16 & .16 & .10 & .10 & .53 & .75 & .73 & -.04 & -.07 & .48 & .26 & .61 & .02 & -.02 & .90 & .33 & .62 & -.23 & -.24 & .75 & .46 & .97 & .17 & .17 & .91 & .06 & .75 & .02 & .24 & .97 \\
 & 1.80 & .17 & .17 & .11 & .11 & .52 & .74 & .73 & -.05 & -.07 & .47 & .27 & .61 & .02 & -.03 & .90 & .30 & .61 & -.26 & -.25 & .76 & .50 & .98 & .21 & .18 & .91 & .06 & .76 & .02 & .26 & .97 \\
 & 1.90 & .18 & .19 & .12 & .13 & .53 & .75 & .73 & -.04 & -.07 & .48 & .27 & .61 & .03 & -.03 & .90 & .33 & .65 & -.23 & -.21 & .76 & .53 & .98 & .24 & .18 & .90 & .07 & .78 & .03 & .28 & .97 \\
 & 2.00 & .19 & .20 & .14 & .14 & .53 & .75 & .73 & -.04 & -.07 & .48 & .28 & .61 & .04 & -.03 & .90 & .18 & .28 & -.38 & -.58 & .72 & .57 & .98 & .28 & .19 & .90 & .09 & .81 & .05 & .30 & .97 \\
\midrule
\multirow[t]{11}{\linewidth}{Bicubic Downscaling} & 1.00 & .06 & .06 & .00 & .00 & .54 & .79 & .80 & .00 & .00 & .51 & .25 & .64 & .00 & .00 & .88 & .56 & .86 & .00 & .00 & .81 & .29 & .80 & .00 & .00 & .84 & .04 & .50 & .00 & .00 & .83 \\
 & 0.95 & .06 & .05 & .00 & -.01 & .51 & .71 & .72 & -.08 & -.08 & .52 & .20 & .52 & -.05 & -.12 & .86 & .58 & .78 & .03 & -.08 & .71 & .29 & .80 & .00 & .00 & .84 & .03 & .43 & -.01 & -.07 & .79 \\
 & 0.90 & .05 & .04 & -.01 & -.02 & .49 & .70 & .71 & -.09 & -.09 & .51 & .19 & .48 & -.06 & -.15 & .84 & .62 & .78 & .07 & -.08 & .67 & .29 & .78 & .00 & -.02 & .83 & .03 & .39 & -.01 & -.11 & .78 \\
 & 0.85 & .05 & .04 & -.01 & -.02 & .49 & .69 & .70 & -.10 & -.10 & .51 & .19 & .46 & -.05 & -.18 & .82 & .64 & .78 & .08 & -.08 & .65 & .27 & .73 & -.01 & -.07 & .80 & .03 & .36 & -.01 & -.14 & .76 \\
 & 0.80 & .04 & .03 & -.01 & -.03 & .48 & .68 & .69 & -.11 & -.10 & .51 & .20 & .44 & -.05 & -.19 & .81 & .62 & .76 & .06 & -.10 & .65 & .27 & .70 & -.02 & -.10 & .78 & .03 & .31 & -.01 & -.19 & .73 \\
 & 0.75 & .04 & .03 & -.02 & -.03 & .47 & .68 & .69 & -.11 & -.11 & .51 & .20 & .43 & -.04 & -.20 & .79 & .67 & .75 & .11 & -.11 & .58 & .26 & .66 & -.03 & -.14 & .76 & .03 & .28 & -.01 & -.23 & .71 \\
 & 0.70 & .03 & .03 & -.02 & -.04 & .48 & .67 & .67 & -.12 & -.12 & .50 & .21 & .42 & -.04 & -.21 & .77 & .67 & .77 & .12 & -.09 & .59 & .25 & .60 & -.04 & -.19 & .73 & .03 & .23 & -.01 & -.28 & .69 \\
 & 0.65 & .03 & .02 & -.03 & -.04 & .47 & .66 & .66 & -.13 & -.14 & .50 & .22 & .42 & -.03 & -.22 & .75 & .69 & .76 & .13 & -.10 & .56 & .24 & .55 & -.05 & -.25 & .70 & .05 & .23 & .01 & -.27 & .65 \\
 & 0.60 & .03 & .02 & -.03 & -.04 & .47 & .66 & .65 & -.14 & -.15 & .49 & .24 & .41 & -.01 & -.22 & .73 & .70 & .77 & .14 & -.09 & .56 & .23 & .48 & -.06 & -.31 & .67 & .05 & .21 & .01 & -.30 & .62 \\
 & 0.55 & .02 & .02 & -.03 & -.04 & .47 & .64 & .63 & -.15 & -.16 & .49 & .25 & .42 & .00 & -.22 & .71 & .70 & .79 & .15 & -.07 & .57 & .22 & .42 & -.07 & -.38 & .64 & .06 & .18 & .02 & -.32 & .60 \\
 & 0.50 & .02 & .01 & -.04 & -.05 & .48 & .63 & .61 & -.16 & -.18 & .49 & .26 & .42 & .01 & -.22 & .70 & .71 & .78 & .15 & -.08 & .57 & .21 & .35 & -.08 & -.45 & .61 & .07 & .16 & .03 & -.34 & .56 \\
\bottomrule
\end{tabular}

\end{table}

\begin{table}[htbp]
    \centering
    \fontsize{5}{5}\selectfont
    \setlength{\tabcolsep}{1.05pt}
    \caption{\textbf{Numerical values of the plots for RAISE-1k (original)\,/\,Synthbuster.}}
    \label{tab:raise1k_vs_synthbuster_plot_values}
    \begin{tabular}{@{\ }p{30pt}rrrrrrrrrrrrrrrrrrrrrrrrrrrrrrr@{\ }}
\toprule
 & & \multicolumn{5}{c}{UnivFD} & \multicolumn{5}{c}{DRCT} & \multicolumn{5}{c}{RINE} & \multicolumn{5}{c}{AIDE} & \multicolumn{5}{c}{SPAI} & \multicolumn{5}{c}{B-Free} \\
 \cmidrule(l{1pt}r{1pt}){3-7} \cmidrule(l{1pt}r{1pt}){8-12} \cmidrule(l{1pt}r{1pt}){13-17} \cmidrule(l{1pt}r{1pt}){18-22} \cmidrule(l{1pt}r{1pt}){23-27} \cmidrule(l{1pt}r{1pt}){28-32}
 & & \multicolumn{2}{c}{Score} & \multicolumn{2}{c}{MSS} & AUC & \multicolumn{2}{c}{Score} & \multicolumn{2}{c}{MSS} & AUC & \multicolumn{2}{c}{Score} & \multicolumn{2}{c}{MSS} & AUC & \multicolumn{2}{c}{Score} & \multicolumn{2}{c}{MSS} & AUC & \multicolumn{2}{c}{Score} & \multicolumn{2}{c}{MSS} & AUC & \multicolumn{2}{c}{Score} & \multicolumn{2}{c}{MSS} & AUC \\
\cmidrule(r{1pt}){3-4} \cmidrule(l{1pt}){5-6}
\cmidrule(r{1pt}){8-9} \cmidrule(l{1pt}){10-11}
\cmidrule(r{1pt}){13-14} \cmidrule(l{1pt}){15-16}
\cmidrule(r{1pt}){18-19} \cmidrule(l{1pt}){20-21}
\cmidrule(r{1pt}){23-24} \cmidrule(l{1pt}){25-26}
\cmidrule(r{1pt}){28-29} \cmidrule(l{1pt}){30-31}
 Transform & Level & Real & Fake & Real & Fake &  & Real & Fake & Real & Fake &  & Real & Fake & Real & Fake &  & Real & Fake & Real & Fake &  & Real & Fake & Real & Fake &  & Real & Fake & Real & Fake &  \\
\midrule
\multirow[t]{11}{*}{JPEG} & - & .33 & .34 & .00 & .00 & .47 & .62 & .84 & .00 & .00 & .81 & .38 & .76 & .00 & .00 & .84 & .61 & .80 & .00 & .00 & .66 & .21 & .87 & .00 & .00 & .93 & .15 & .93 & .00 & .00 & .98 \\
 & 100 & .33 & .30 & .01 & -.05 & .43 & .82 & .86 & .19 & .02 & .65 & .44 & .71 & .06 & -.05 & .76 & .21 & .50 & -.40 & -.30 & .73 & .24 & .87 & .03 & .01 & .92 & .15 & .94 & .01 & .01 & .98 \\
 & 90 & .28 & .22 & -.05 & -.12 & .38 & .56 & .68 & -.07 & -.17 & .65 & .31 & .51 & -.07 & -.25 & .72 & .10 & .14 & -.51 & -.66 & .54 & .19 & .88 & -.01 & .01 & .94 & .15 & .93 & .00 & .00 & .98 \\
 & 80 & .28 & .23 & -.05 & -.11 & .40 & .74 & .80 & .12 & -.05 & .62 & .31 & .54 & -.07 & -.21 & .79 & .12 & .19 & -.49 & -.61 & .58 & .19 & .86 & -.02 & .00 & .93 & .14 & .91 & -.01 & -.02 & .98 \\
 & 70 & .31 & .24 & -.01 & -.11 & .38 & .82 & .85 & .20 & .00 & .60 & .36 & .65 & -.02 & -.11 & .85 & .15 & .22 & -.47 & -.58 & .59 & .18 & .81 & -.02 & -.05 & .91 & .14 & .90 & .00 & -.03 & .97 \\
 & 60 & .28 & .23 & -.05 & -.12 & .40 & .77 & .82 & .15 & -.03 & .62 & .37 & .62 & -.01 & -.14 & .85 & .16 & .22 & -.46 & -.57 & .58 & .16 & .74 & -.04 & -.12 & .89 & .16 & .88 & .01 & -.05 & .96 \\
 & 50 & .28 & .23 & -.05 & -.11 & .41 & .75 & .76 & .13 & -.08 & .55 & .39 & .62 & .01 & -.14 & .85 & .17 & .23 & -.44 & -.57 & .56 & .16 & .71 & -.05 & -.16 & .87 & .17 & .86 & .02 & -.07 & .95 \\
 & 40 & .32 & .24 & -.01 & -.10 & .39 & .77 & .72 & .15 & -.12 & .47 & .40 & .59 & .02 & -.17 & .83 & .19 & .27 & -.43 & -.52 & .60 & .12 & .63 & -.08 & -.23 & .86 & .19 & .84 & .04 & -.09 & .93 \\
 & 30 & .38 & .25 & .05 & -.10 & .36 & .77 & .70 & .15 & -.15 & .44 & .42 & .61 & .04 & -.15 & .84 & .20 & .31 & -.42 & -.49 & .62 & .09 & .53 & -.11 & -.33 & .85 & .21 & .80 & .06 & -.13 & .90 \\
 & 20 & .42 & .29 & .09 & -.06 & .38 & .84 & .76 & .21 & -.08 & .43 & .44 & .62 & .05 & -.13 & .85 & .20 & .29 & -.41 & -.50 & .60 & .06 & .35 & -.14 & -.52 & .76 & .24 & .73 & .09 & -.20 & .85 \\
 & 10 & .60 & .48 & .27 & .14 & .43 & .85 & .84 & .23 & -.01 & .50 & .47 & .63 & .09 & -.13 & .82 & .21 & .32 & -.40 & -.47 & .64 & .00 & .05 & -.21 & -.82 & .71 & .31 & .59 & .17 & -.34 & .72 \\
\midrule
\multirow[t]{11}{\linewidth}{Bilinear Upscaling} & 1.00 & .33 & .34 & .00 & .00 & .47 & .62 & .84 & .00 & .00 & .81 & .38 & .76 & .00 & .00 & .84 & .61 & .80 & .00 & .00 & .66 & .21 & .87 & .00 & .00 & .93 & .15 & .93 & .00 & .00 & .98 \\
 & 1.10 & .32 & .33 & .00 & -.01 & .46 & .56 & .84 & -.07 & -.01 & .84 & .33 & .66 & -.05 & -.10 & .85 & .53 & .65 & -.08 & -.14 & .61 & .27 & .93 & .07 & .06 & .94 & .12 & .92 & -.03 & -.01 & .98 \\
 & 1.20 & .34 & .35 & .02 & .01 & .47 & .57 & .83 & -.05 & -.01 & .83 & .34 & .62 & -.05 & -.14 & .83 & .58 & .72 & -.03 & -.07 & .63 & .27 & .93 & .06 & .07 & .95 & .14 & .94 & -.01 & .01 & .99 \\
 & 1.30 & .36 & .38 & .03 & .04 & .48 & .59 & .85 & -.03 & .00 & .84 & .35 & .62 & -.03 & -.14 & .83 & .39 & .41 & -.22 & -.39 & .52 & .27 & .94 & .06 & .07 & .95 & .16 & .95 & .01 & .02 & .99 \\
 & 1.40 & .38 & .40 & .05 & .06 & .49 & .61 & .88 & -.01 & .04 & .87 & .37 & .68 & -.01 & -.08 & .87 & .50 & .61 & -.11 & -.18 & .61 & .25 & .93 & .05 & .07 & .95 & .17 & .95 & .02 & .02 & .99 \\
 & 1.50 & .38 & .41 & .05 & .06 & .50 & .60 & .87 & -.02 & .03 & .87 & .37 & .66 & -.01 & -.10 & .86 & .64 & .75 & .02 & -.05 & .64 & .30 & .94 & .10 & .08 & .95 & .18 & .95 & .03 & .02 & .99 \\
 & 1.60 & .40 & .42 & .07 & .08 & .50 & .60 & .87 & -.02 & .02 & .86 & .38 & .66 & .00 & -.10 & .86 & .42 & .34 & -.19 & -.46 & .45 & .31 & .94 & .11 & .07 & .94 & .19 & .95 & .04 & .02 & .99 \\
 & 1.70 & .40 & .44 & .08 & .10 & .51 & .61 & .86 & -.01 & .02 & .85 & .39 & .65 & .01 & -.11 & .85 & .51 & .58 & -.11 & -.21 & .58 & .34 & .93 & .14 & .06 & .93 & .20 & .95 & .06 & .02 & .99 \\
 & 1.80 & .42 & .46 & .09 & .12 & .52 & .61 & .85 & -.01 & .01 & .84 & .39 & .63 & .01 & -.12 & .84 & .50 & .53 & -.11 & -.27 & .53 & .38 & .92 & .17 & .06 & .92 & .21 & .94 & .07 & .01 & .99 \\
 & 1.90 & .43 & .48 & .10 & .13 & .53 & .62 & .85 & -.01 & .00 & .84 & .40 & .64 & .02 & -.12 & .85 & .58 & .59 & -.04 & -.21 & .54 & .43 & .92 & .23 & .06 & .90 & .22 & .94 & .08 & .01 & .99 \\
 & 2.00 & .43 & .47 & .10 & .12 & .52 & .59 & .84 & -.03 & -.01 & .84 & .40 & .63 & .02 & -.13 & .83 & .39 & .29 & -.22 & -.50 & .47 & .60 & .96 & .40 & .10 & .88 & .23 & .93 & .09 & .00 & .98 \\
\midrule
\multirow[t]{11}{\linewidth}{Bilinear Downscaling} & 1.00 & .33 & .34 & .00 & .00 & .47 & .62 & .84 & .00 & .00 & .81 & .38 & .76 & .00 & .00 & .84 & .61 & .80 & .00 & .00 & .66 & .21 & .87 & .00 & .00 & .93 & .15 & .93 & .00 & .00 & .98 \\
 & 0.95 & .28 & .28 & -.05 & -.06 & .45 & .53 & .82 & -.09 & -.02 & .84 & .32 & .68 & -.06 & -.08 & .85 & .60 & .81 & -.02 & .01 & .69 & .30 & .91 & .10 & .04 & .92 & .10 & .91 & -.05 & -.02 & .98 \\
 & 0.90 & .26 & .24 & -.07 & -.10 & .42 & .51 & .83 & -.11 & -.01 & .86 & .30 & .65 & -.08 & -.11 & .84 & .59 & .83 & -.02 & .03 & .71 & .31 & .90 & .11 & .04 & .91 & .09 & .90 & -.05 & -.04 & .98 \\
 & 0.85 & .24 & .25 & -.09 & -.10 & .44 & .51 & .83 & -.12 & -.01 & .86 & .29 & .64 & -.09 & -.12 & .85 & .60 & .83 & -.02 & .03 & .71 & .35 & .88 & .15 & .02 & .89 & .09 & .88 & -.06 & -.05 & .97 \\
 & 0.80 & .23 & .22 & -.10 & -.12 & .43 & .50 & .84 & -.12 & .00 & .87 & .28 & .64 & -.10 & -.12 & .86 & .59 & .83 & -.03 & .04 & .72 & .34 & .87 & .13 & .00 & .88 & .08 & .86 & -.07 & -.07 & .96 \\
 & 0.75 & .20 & .20 & -.13 & -.14 & .42 & .49 & .85 & -.14 & .01 & .89 & .27 & .63 & -.11 & -.13 & .87 & .58 & .82 & -.03 & .02 & .72 & .36 & .84 & .15 & -.02 & .86 & .07 & .84 & -.08 & -.09 & .95 \\
 & 0.70 & .19 & .19 & -.14 & -.15 & .42 & .47 & .85 & -.15 & .01 & .90 & .27 & .63 & -.11 & -.12 & .87 & .57 & .83 & -.04 & .04 & .74 & .35 & .82 & .14 & -.05 & .84 & .06 & .81 & -.08 & -.12 & .94 \\
 & 0.65 & .18 & .17 & -.15 & -.17 & .41 & .48 & .86 & -.15 & .01 & .90 & .26 & .63 & -.12 & -.12 & .88 & .58 & .83 & -.03 & .03 & .72 & .34 & .77 & .14 & -.09 & .81 & .06 & .78 & -.09 & -.15 & .92 \\
 & 0.60 & .17 & .16 & -.16 & -.18 & .41 & .48 & .86 & -.15 & .01 & .90 & .27 & .64 & -.12 & -.11 & .88 & .59 & .83 & -.02 & .03 & .73 & .32 & .73 & .12 & -.14 & .79 & .05 & .74 & -.09 & -.19 & .90 \\
 & 0.55 & .15 & .15 & -.18 & -.20 & .41 & .46 & .85 & -.16 & .00 & .90 & .25 & .63 & -.13 & -.13 & .88 & .58 & .81 & -.03 & .02 & .72 & .33 & .69 & .13 & -.18 & .75 & .05 & .70 & -.10 & -.23 & .89 \\
 & 0.50 & .14 & .13 & -.19 & -.21 & .42 & .45 & .82 & -.17 & -.02 & .88 & .24 & .59 & -.14 & -.17 & .88 & .58 & .74 & -.04 & -.06 & .65 & .33 & .63 & .12 & -.23 & .71 & .04 & .65 & -.11 & -.28 & .87 \\
\bottomrule
\end{tabular}

\end{table}

\end{document}